%% file: ijcai23.tex
\documentclass{article}
\usepackage{ijcai23}

\usepackage{times}
\usepackage{soul}
\usepackage{url}
\usepackage[hidelinks]{hyperref}
\usepackage[utf8]{inputenc}
\usepackage[small]{caption}
\usepackage{graphicx}
\usepackage[graphicx]{realboxes}
\usepackage{amsmath}
\usepackage{amsthm}
\usepackage{booktabs}
\usepackage{algorithm}
\usepackage{algorithmic}
\usepackage[switch]{lineno}

\usepackage{latexsym}

\newcommand{\ifshowsupplementary}{\iftrue}

\title{Domain-Adaptive Self-Supervised Face \& Body Detection in Drawings}

\author{Anonymous Authors}

\author{
Barış Batuhan Topal$^1$ \and
Deniz Yuret$^1$ \and
Tevfik Metin Sezgin$^1$ \\
\affiliations
$^1$Department of Computer Engineering, KUIS AI Center, Koç University\\
\emails
\{baristopal20, dyuret, mtsezgin\}@ku.edu.tr,
}

\begin{document}

\maketitle

\begin{abstract}
    Drawings are powerful means of pictorial abstraction and communication. Understanding diverse forms of drawings, including digital arts, cartoons, and comics, has been a major problem of interest for the computer vision and computer graphics communities. Although there are large amounts of digitized drawings from comic books and cartoons, they contain vast stylistic variations, which necessitate expensive manual labeling for training domain-specific recognizers. In this work, we show how self-supervised learning, based on a teacher-student network with a modified student network update design, can be used to build face and body detectors. Our setup allows exploiting large amounts of unlabeled data from the target domain when labels are provided for only a small subset of it. We further demonstrate that style transfer can be incorporated into our learning pipeline to bootstrap detectors using a vast amount of out-of-domain labeled images from natural images (i.e., images from the real world). Our combined architecture yields detectors with state-of-the-art (SOTA) and near-SOTA performance using minimal annotation effort. Our code can be accessed from {\fontfamily{qcr}\selectfont \url{https://github.com/barisbatuhan/DASS_Detector}}.
\end{abstract}

\section{Introduction}

Drawings serve as a rich and expressive medium for communication. The earliest examples were painted on cave walls more than 45,000 years ago \cite{brumm2021oldest}. Here we focus on comic books and cartoons, which are relatively recent forms of media. They combine text and graphics in a unique format to convey narratives. Key problems such as extracting the visual structure of the scenes, understanding the accompanying text, and modeling how they connect to form the narrative pose significant challenges. Hence, understanding comics has been a problem of interest to computer vision, computer graphics, and NLP communities.

\begin{figure}
    \centering
    \includegraphics[width=0.36\linewidth]{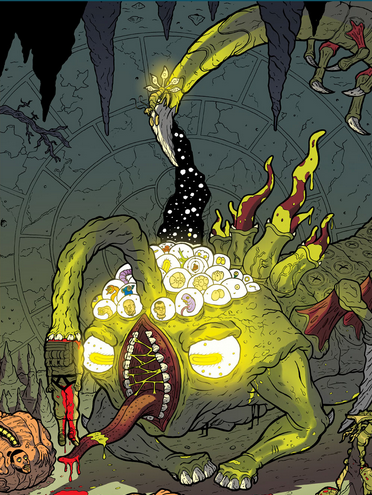}
    \includegraphics[width=0.59\linewidth]{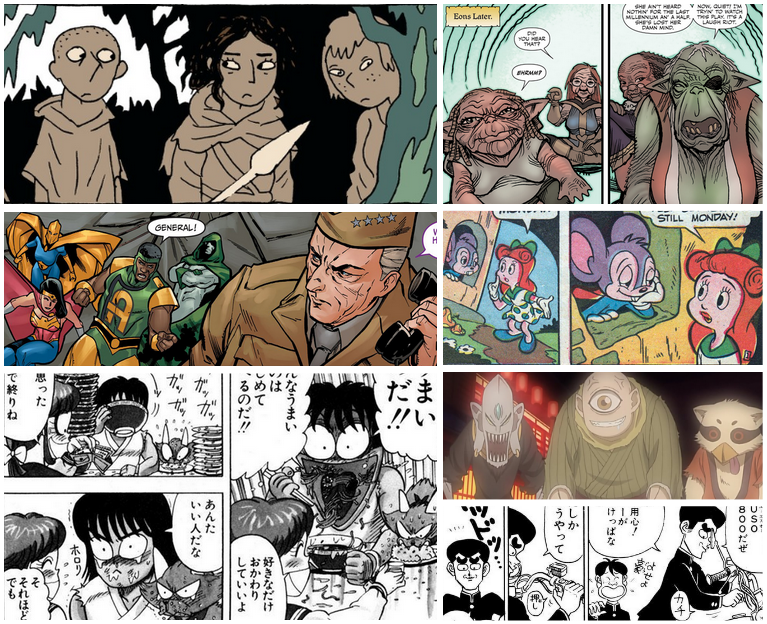}
    \caption{Examples on the adversity of this domain (left: non-human character, right: samples from different character designs and styles).}
    \label{fig:adversity}
\end{figure}

In drawings, the story is narrated primarily through the scene's main characters. Hence, we study on face and body detection, two primary problems for understanding drawings. Training face and body detectors is complicated by two challenges. First, although a tremendous amount of unlabeled data is available (primarily as digitized comic book pages and animations), face and body annotations are largely lacking. Second, since character design and drawing style change substantially across artists, series, and cultures (see Figure \ref{fig:adversity}), each domain inevitably requires domain-specific tuning to create detectors. In this work, we present a pre-training pipeline for creating domain-adapted detectors, which addresses both problems. Our pipeline has two major components. The first is a self-learning component that can exploit vast amounts of unlabeled data from the target domain to create detectors that can be tuned with minimal labeled data. More specifically, we introduce a modified version of teacher-student architecture to drawings, where we periodically update the student network's weights with teacher's after a specific number of iterations and utilize the OHEM \cite{shrivastava2016training} loss with an additional positive and negative confidence threshold limitation for a more stable training.  We show that this self-learning model works best if it starts with a sufficiently good teacher. This component leads to the second key component of our pipeline, which uses style transfer to transform vast amounts of labeled natural images to create sufficiently good teacher models by utilizing 11 styles from 4 style transfer algorithms.

\begin{figure*}[ht]
    \centering
    \includegraphics[width=0.96\textwidth]{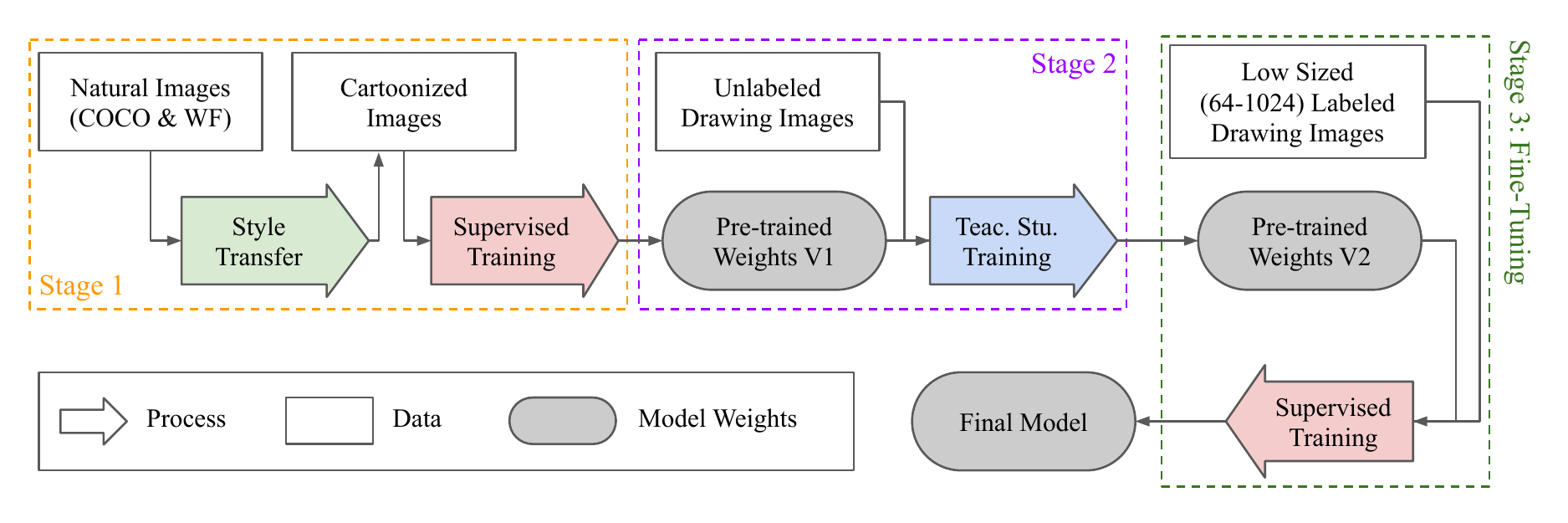}
    \caption{Summary of the proposed pipeline.}
    \label{fig:complete_pipeline}
\end{figure*}

We employ a multi-tasking strategy by jointly training the model for faces and bodies to reduce inference time and to benefit from the contextual and spatial relationship. To utilize datasets with face-only and body-only annotations, we use two detection heads: one to predict the faces, and the other for bodies. Even without drawing domain supervision, our teacher-student model outperforms previous supervised SOTA of DCM 772 \cite{jimaging4070089} and weakly-supervised SOTA \cite{inoue2018crossdomain} in most datasets. When initialized with our pre-trained weights, our supervised model sets a new SOTA performance for most datasets, even if limited drawing data is used in training.

\section{Related Works}
\label{section:rel_works}

\subsection{Detection}

With the increasing size of annotated data, models with high dependence on supervision were able to get good results (e.g., \cite{bochkovskiy2020yolov4,DynamicRCNN,yolox2021}). \cite{liu2021unbiased} and \cite{xu2021end} introduced teacher-student training schemes and gained a significant performance boost with a low amount of labeled data. Unlike this work, these studies target natural images. Thus, cross-domain detection with these models is prone to false positives (FP) and negatives (FN). Several studies have improved the teacher-student scheme to work well in cross-domain detection. While MTOR \cite{mtor_model} exploits object relations in region-level consistency, inter-graph consistency, and intra-graph consistency, UMT \cite{umt_model} tries to eliminate teacher and student network biases through distillation and style transferring, D-adapt \cite{dadapt} adopts an adversarial pipeline to the detector model, H$^2$FA R-CNN \cite{h2farcnn} utilizes weak supervision and domain classifiers to create a more domain-invariant model. Although our solution is more similar to UMT compared to other cross-domain studies, we improve its style transferring part by mixing multiple styles, we modify the standard teacher-student training to compensate for the FP and FN cases, and we change the loss function to force the model to learn from more confident proposals. 

Several studies have been done on face and object detection, specifically in drawings.  \cite{zhang2020acfd} proposed a fully-supervised face detector using only iCartoonFace; \cite{ogawa2018object} trained a detector from Manga 109; \cite{jimaging4070089} used DCM 772; \cite{inoue2018crossdomain} utilized Comic2k, Watercolor2k, and Clipart1k. However, these models are only trained on specific sub-domains of drawings (i.e., only utilized a single dataset with limited stylistic coverage). In this study, we leverage unlabeled drawing images from any sub-drawing domain and show that the performance on drawings can be significantly improved by using an effective pre-training pipeline and a better detector architecture.

\subsection{Style Transfer}
\label{subsection:style_trans}

 Conversion of natural images to drawings is an unpaired image-to-image translation task. SOTA models for this task have been designed with U-Net-like Generative Adversarial Networks (i.e., down-sampling first and then up-sampling). We use several cartoonization models to increase the stylistic variety of the pre-training data by selecting 11 styles from these works: Monet, Van Gogh, Cezanne from CycleGAN \cite{CycleGAN2017}; Shinkai, Hayao, Hosoda, Paprika from CartoonGAN \cite{chen2018cartoongan}; AS, KH, Miyazaki from GANILLA \cite{DBLP:journals/corr/abs-2002-05638}; and the default style in White-Box Cartoonization \cite{Wang_2020_CVPR}. While previous detection studies on drawings have also utilized style transfer methods (e.g., \cite{inoue2018crossdomain,umt_model}), we improve on these results by combining multiple styles and analyzing which styles increase the performance more.

\subsection{Datasets}
\label{subsection:datasets}

Digitization has made millions of unlabeled drawings reachable on the internet. Thousands of old comic book series (e.g., Golden Age Comics between the 1930s - 1950s) have been published on several websites \footnote{comicbookplus.com and digitalcomicmuseum.com} and gathered as an unlabeled dataset named COMICS \cite{iyyer2017amazing}. Newer series can be obtained through web crawling. Unfortunately, annotated datasets only comprise a small subset of this domain in terms of stylistic variety and quantity. Regarding the stylistic distribution of labeled datasets, the majority of iCartoonFace \cite{zheng2020cartoon} is retrieved from Asian products ($\sim$74\%), Manga 109 \cite{mtap_matsui_2017} only covers Japanese Manga styles, DCM 772 \cite{jimaging4070089} is limited to comics from Golden Age Era. Although \cite{inoue2018crossdomain} introduces Comic2k, Watercolor2k, and Clipart1k for body detection, they also remain stylistically bound in their sub-domain. Currently, none of the available datasets provide comprehensive stylistic coverage. In particular, contemporary US and Western comics have little if any annotated examples. In terms of dataset quantity, iCartoonFace contains a significant amount of face data with its 50,000 training and 10,000 validation images. The situation is a bit more challenging with body annotations: Manga 109 has $\sim$21,000 page images, but the style is limited to black and white mangas. DCM 772 consists of only 772 images. Comic2k, Watercolor2k, and Clipart1k increase the total labeled data by 2,500 instances. Building a body detector for drawings that is not fragile to different styles is challenging using only these datasets. Hence, self-supervised approaches are essential for creating suitable models for target instances with unseen styles.

\begin{figure*}[ht]
    \centering
    \includegraphics[width=0.95\textwidth]{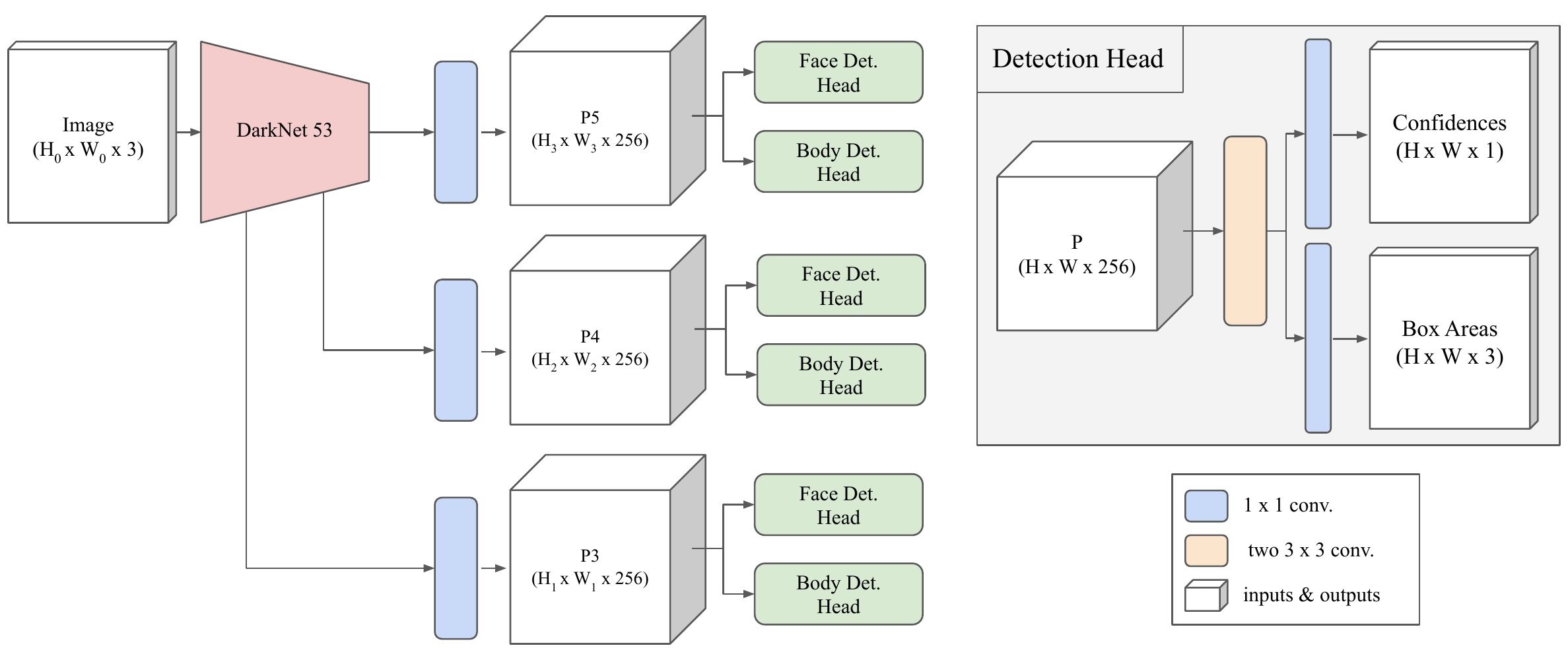}
    \caption{Our complete model architecture.}
    \label{fig:model_arch}
\end{figure*}

\section{Methodology}
\label{section:methodology}

Our training consists of three stages. In the first stage, we use two large and annotated real-life image datasets, cartoonize them using style transfer methods, and perform pre-training for face and body detection. In the second stage, we utilize the extensive amount of unlabeled comic drawings available and perform self-supervised training on our pre-trained model with the modified form of the teacher-student architecture. In the final stage, we leverage the limited amount of annotated comic drawings to fine-tune our model. In Figure \ref{fig:complete_pipeline}, you can see a demonstration of our complete pipeline. In the following subsections, we describe our base model and the three stages we propose in more detail.

\subsection{Model Architecture}
\label{subsection:model_arch}

Since the challenge in our domain consists of stylistic variety in object representations (see Figure \ref{fig:adversity}), we decide that adopting an object-detector-like model would provide greater performance, where the architecture is specifically designed to find multiple objects with various appearances. Secondly, we aim to use a more robust and simple model with low inference time to focus mainly on the effects of style transfer and self-supervised training. Therefore, we select one of the SOTA single-shot non-swin-transformer anchor-free object detectors, YOLOX \cite{yolox2021}, as our baseline architecture. However, our pre-training pipeline does not depend on this specific baseline. Hence it can be applied to any detector.

As discussed in Section \ref{section:rel_works}, COCO \cite{lin2014microsoft}, WIDER FACE \cite{yang2016wider}, and some of the available drawing datasets do not include both face and body annotations together. To train the model jointly for both face and body parts and benefit from all the available datasets, we separate the detection head of the original YOLOX model into two pieces. Each piece proposes bounding boxes with their confidence values only for a single class. Our overall architecture can be seen in Figure \ref{fig:model_arch}. During training, the heads are trained alternately at each forward pass.

\subsection{Stage 1: Style Transferred Pre-Training}

\paragraph{Preprocessing.}{We process COCO and WIDER FACE datasets with 11 different styles. We eliminate all the images in COCO that do not have people or animals. We also count animals as bodies during training because drawings may include animal-like characters. To the best of our knowledge, no dataset includes annotations for animal faces. Thus, facial training is solely done through human faces in WIDER FACE. We discard the images in which a person has a face with its maximum facial side length smaller than $\sim$2\% of the image's minimum side length. These faces are not required in the dataset since characters in drawings mostly have a bigger appearance on the image.}

\paragraph{Training Experiments.}{We create 5 different experiments to test our model's success at pre-training stage 1:

\begin{itemize}
    \item \textbf{Single Styles:} We analyze the effect of each style on the detection performance by training individual models with only one style transferring method.
    \item \textbf{All Styles:} We train an additional model by combining all styles with random selection per each image to notice if using multiple styles increases the overall performance.
    \item \textbf{Best Styles:} We choose five styles that result in the greatest performance individually and train another model by combining only these to find if selecting the most effective styles is more logical instead of utilizing all styles.
    \item \textbf{No Style:} We train an extra model that uses the original images without any stylization to observe the benefit of style transferring.
    \item \textbf{No Animals Included:} We test the effect of including animal bodies to body annotations to the performance. We utilize all of the styles but exclude the animal boxes from the training data.
\end{itemize}
}

\subsection{Stage 2: Self-Supervised Pre-Training}
\label{subsection:teac_stu}

\begin{figure*}
    \centering
    \includegraphics[width=0.9\textwidth]{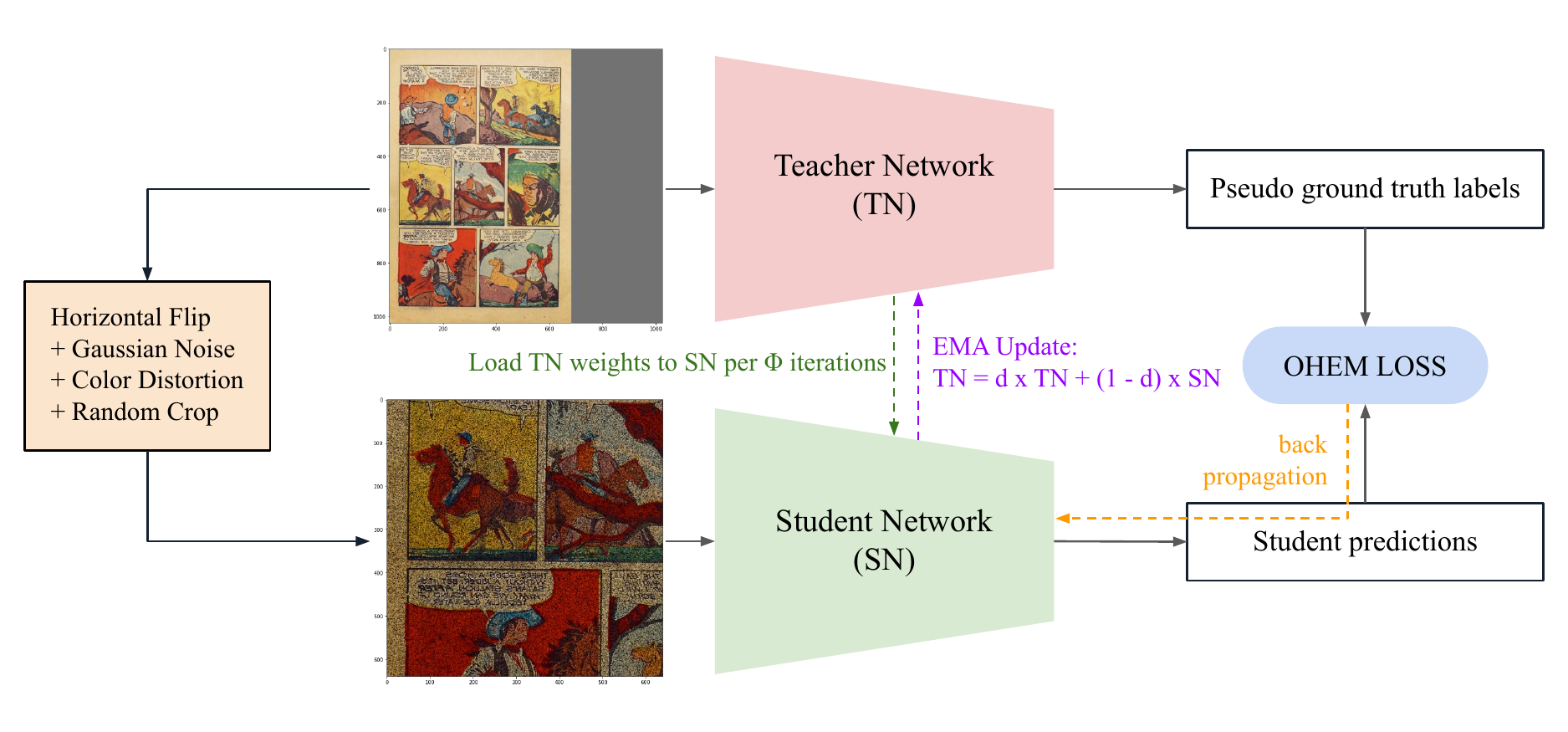}
    \caption{Our stage 2 teacher-student network training process.}
    \label{fig:uns_arch}
\end{figure*}

\paragraph{Model Architecture.}{The model consists of two different network parts: teacher and student. These networks are identical and initialized from the same pre-trained set of weights that we obtain from stage 1: style transferred pre-training with the mixture of all styles. The teacher network processes a non-augmented complete image and generates bounding box predictions along with their confidence values. The student network also generates predictions, but it processes a heavily augmented version of the same input image. High-confidence predictions of the teacher network are further processed with the non-maximum suppression (NMS) algorithm, and the outputs are considered as the pseudo-ground-truth labels of the image. The student network is trained with the loss computed by using the labels retrieved from the teacher network. The gradient flow of the teacher network is stopped, and it is updated at each iteration with respect to the Eq. \ref{eq:tn_update}, where $TN$ is the teacher, $SN$ is the student network weight, $d$ is a hyper-parameter:

\begin{equation}
\begin{split}
    TN = d \cdot TN + (1 - d) \cdot SN
\end{split}
\label{eq:tn_update}
\end{equation}

\noindent Although TN is updated with the student weights in earlier studies, student weights are only changed with backpropagation. In our experiments, we have seen that this design causes the development of both modules at the earlier stages but a significant performance drop in SN in the later iterations due to the noisy pseudo-ground-truth labels caused by the change in the input domain between pre-training stage 1 (i.e., cartoonized natural images) and self-supervised processes (i.e., drawings). This drop also affects the performance of TN. Hence, we load the weights of TN to SN per each $\Phi$ iteration to fix the deterioration of SN. Since this step manipulates the values without the gradient flow, an optimizer with the momentum information may mislead the overall model. Thus, we change our optimizer to Stochastic Gradient Descent (SGD). Our self-supervised architecture can be seen in Figure \ref{fig:uns_arch}.}

\paragraph{Loss.}{In Focal Loss, each prediction is included in the confidence loss calculation with a weight that balances the positive (i.e., predictions in which the actual ground truth object is present) and negative (i.e., predictions that point to a background area) boxes. This approach is advantageous in fully supervised training since the ground truth box areas of every object in the image are given to the model. On the other hand, in self-supervised detectors, the high-probability predictions of the teacher model are selected as pseudo-ground-truth values, which are prone to false positives (FP) and false negatives (FN). FP cases can be minimized by increasing the confidence threshold for ground truth selection. However, this choice also increases the FN rate. To further decrease the FN cases, we follow the OHEM loss \cite{shrivastava2016training}, where only a subset of predictions are chosen to calculate the loss. We also modify this loss so that the predictions can be selected as positive predictions only above a specific confidence threshold and negative predictions below a particular threshold. Subset selection and this modification help the model to skip a subset of FN cases of the teacher model in loss calculations (e.g., if a face/body area is predicted but has a low confidence value). Loss calculation of a single selected box proposal can be seen in Eq. \ref{eq:modif_ohem}:

\begin{equation}
\begin{split}
&\mathcal{L}_{conf} = - p \cdot ct_{pos} \cdot log(\hat{p}) - (1-p) \cdot ct_{neg} \cdot log(1 - \hat{p}) \\
&\mathcal{L}_{reg} = \sum_{i}^{\{w, h, x, y\}} smooth_{L_1}(i_{gt}, i_{pred}) \\
&\mathcal{L}_{total} = \mathcal{L}_{conf} + \beta \mathcal{L}_{reg} 
\end{split}
\label{eq:modif_ohem}
\end{equation}

\noindent $\mathcal{L}_{conf}$ is the confidence loss and $\mathcal{L}_{reg}$ is the regression loss. $p \in \{0, 1\}$ indicates if the box is selected as positive ($p = 1$) or negative ($p=0$), $\hat{p} \in [0, 1]$ is the confidence value of the selected box, $ct_{pos} \in \{0, 1\}$ is 1 if the confidence of the proposed box is above the positive confidence threshold, $ct_{neg} \in \{0, 1\}$ is 1 if the confidence of the proposed box is below the negative confidence threshold, $\{w, h, x, y\}$ are the width, height, and the center points of the box, $\beta$ is the balancing parameter between confidence and regression losses.}

\paragraph{Unlabeled Datasets.}{We crawled 195,321 comic book pages from today's US and European series to train our model. We also utilized 198,657 pages from COMICS and leveraged iCartoonFace, Manga 109 pages, Comic2k, Watercolor2k, and Clipart1k images. At each forward pass, we select a random image from these image sets.}

\paragraph{Experiments \& Hyper-parameters.}{We run several experiments with different losses, $\Phi$, $\beta$, $d$, positive and negative student confidence thresholds. In our final model, we set $\Phi$ to 500, $\beta$ to 2, $d$ to 0.9996, and positive and negative thresholds ($ct_{pos}^{thres}$ and $ct_{neg}^{thres}$) to 0.5.}

\subsection{Stage 3: Fine-Tuning}

We conduct experiments with three different pre-training methods: random initialization, style transferred pre-training in stage 1, and teacher-student network from stage 2. Since each drawing dataset contains its own separate stylistic distribution, they should be fine-tuned separately to obtain the maximum performance on their test set. Thus, we fine-tune the model with single datasets for each pre-training variation by randomly selecting a limited number of image instances (i.e., 64, 128, 256, 512, 1024 images, or all data). As Manga 109 and DCM 772 consist of page images instead of individual panels, we separate panels during training to increase the number of input data and test the models with their page images.

\section{Results \& Discussion}
\label{section:res_exp}

In the following parts, we explain our training details, discuss the effect of style transferring in stage 1, analyze the experiments done by utilizing the teacher-student network, and present our results retrieved after fine-tuning with limited and unlimited drawing data. We will use abbreviations\footnote{iCartoonFace as iCF, Manga 109 faces as M109-F,  Manga 109 bodies as M109-B,  DCM 772 faces as DCM-F, DCM772 bodies as DCM-B, Comic2k as C2k, Watercolor2k as W2k, and Clipart1k as C1k. If Manga 109 is used directly, then it means that the face and body AP scores are averaged.} of datasets in the given tables to save space since there are many datasets for evaluation. Average Precision (AP) is selected as the evaluation metric for detection, and the intersection of union value for evaluation is fixed at 0.5. At each table given in this section, the best result per column is marked in \textbf{bold} and the second is \underline{underlined}.

\subsection{Training Details}

In all variations and experiments, the batch size is set to 16, and one Tesla T4 GPU is used. AP scores are calculated by running the same variation five times and computing the average of these runs. At stages 1 and 3, the learning rate is fixed at 0.001. The highest-scoring checkpoints in the evaluation set among 350 epochs are chosen as the final models. The first and the last 15 epochs include no augmentation. Otherwise, horizontal \& vertical flips, the color distortion between $[-20^\circ, 20^\circ]$ degrees, shear, and mosaic augmentation (i.e., combining four random images and passing them as a single image) are applied randomly between the 15th and 335th epochs. For the teacher-student network, the learning rate is set as 0.0001, and the best checkpoints in 10000 iterations are taken as final models. While the input image of the teacher network is only horizontally flipped, Gaussian noise, color distortion, and random crop are applied additionally to the student network in all epochs.

\input{tables/stage_2_results}

\input{tables/stage_1_results}

\subsection{Stage 1: Style Transferred Pre-Training}
\label{res:sty_train}

In this stage, we try to find the best combination to initialize the teacher-student network. For this purpose, we train the model variations with cartoonized natural images but evaluate them with drawing datasets. Scores retrieved after pre-training stage 1 are given in Table \ref{table:stage1_results} for the individual top-5 styles (i.e., \textit{Whitebox, Hosoda, KH, Hayao} and \textit{Shinkai}) and other experiments.

In the drawing domain, characters can be drawn in various styles. Although texture and colors continuously change among products, key fragments of faces and bodies preserve their existence (e.g., faces include at least one eye, and bodies contain either a head, arms, or legs). In our case, we believe that using multiple styles instead of one forces the model to focus more on to shape of the object rather than texture. Consequently, the model learns more generalizable information rather than style-specific; the objects are detected more accurately when the model is tested with unseen examples. Therefore, while leveraging even a single style transferring method from top-5 ensures performance increase compared to using \textit{No Styles}, \textit{All Styles} outperforms both individual styles and \textit{Top-5 Styles}. Furthermore, adding animal annotations to the ground truth during the style transferred pre-training stage pushes the performance even further.

\subsection{Stage 2: Self-supervised Pre-Training}
\label{res:selfsup_train}

In this Section, we discuss all of our experiments in the self-supervised stage. We will refer to Table \ref{table:uns_exps_params} for the additional student network (SN) update interval ($\Phi$), loss selection, positive ($c_{pos}^{thold}$) and negative ($c_{neg}^{thold}$) SN confidence thresholds, usage of momentum in the optimizer ($\gamma$), and for highlighting the importance of style transferring before the self-supervised stage (ST). \\

\paragraph{Loss.}{In experiments 2, 4, 5, and 6, our modified OHEM loss is compared with the SimOTA loss, which is the default loss method in YOLOX and an advanced variation of Focal loss. We believe that selecting a subset of predictions for backpropagation reduces the amount of misleading in FP and FN cases. Our results also validate that OHEM loss is more suitable for our self-supervised architecture. Models with OHEM loss outperform others with up to $\sim2.8$ AP difference.}

\paragraph{Updating SN per $\Phi$ Iterations.}{Between experiments 1 and 4, we try various iteration counts for $\Phi$. We observe that the overall performance drops if $\Phi > 500$. The score is worst if there is no manual SN update (i.e., $\Phi = None$).}


\input{tables/stage_3_results.tex}

\paragraph{Student Confidence Thresholds ($c_{pos}^{thold}$ and $c_{neg}^{thold}$).}{We test the influence of positive and negative SN confidence thresholds in experiments 2 and 7-10. With a threshold starting from too high for positive and too low for negative (exp. 7), the average performance is lower than the others. While the original OHEM loss corresponds to exp. 10, adding additional thresholds for SN results in greater or similar scores (e.g., experiments 8 and 9). The best performance is obtained by setting both $c_{pos}^{thold}$ and $c_{neg}^{thold}$ to 0.5.}

\paragraph{Optimizer Selection.}{Our study states that manually changing SN's weights with TN's may mislead the overall model if an optimizer with momentum is utilized. To test our statement, we train two models with the same hyper-parameter configurations but select standard SGD in one and Nesterov SGD in the other (exp. 2 and 12). In almost every dataset, standard SGD scores $\sim1.5-2\%$ higher.}

\paragraph{Style Transferring Before Self-supervised Stage 2.}{We investigate if style transferring is needed in stage 1 before applying self-supervised stage 2. We train two models with the same settings but initialize the pre-trained weights of these models in the teacher-student stage differently: one with the weights retrieved from pre-training stage 1, including style transferring, the other without style transferring (exp. 2 and 11). Overall, AP difference is $\sim7\%$. Hence, applying style transferring in stage 1 has a significant positive effect on the self-supervised stage 2 model performance.}

\subsection{Stage 3: Fine-Tuning}
\label{res:sup_train}

\input{tables/pretrain_comparisons.tex}

We train our architecture for single datasets with limited instances to evaluate their behavior when only a low amount of data is available. The average of scores for all datasets (i.e., iCartoonFace, Manga 109, DCM 772, Comic2k, Watercolor2, Clipart 1k) are shared in Table \ref{table:single_best_avg}. In the cases with extremely low instance counts (i.e., 64 and 512 images), utilization of natural images and self-supervised learning results in up to $\sim24\%$ performance increase compared to starting from a random initial state. When trained with all available data, both style-transferring-based and teacher-student-based pre-training methods score similar values. We believe this is caused since there is sufficient data for these specific sub-domains to close the gap that emerged from the self-supervised stage. However, we still obtain a significant improvement ($\sim2.2\%$) when models start from pre-trained weights instead of random initialization. This shows that leveraging style transferred pre-training enhances the performance independently from the amount of labeled fine-tuning data. 

In Table \ref{table:models_perf}, we compare previous SOTA models with our results from each stage checkpoint (i.e., style transfer, teacher-student, fine-tuning with individual datasets). Our model achieves close scores to ACFD \cite{zhang2020acfd} and outperforms other SOTA models. Even with a low amount of training images, we obtain better or comparable results with \cite{jimaging4070089} and \cite{ogawa2018object}. Increasing the model size from the tiny version of YOLOX to the XL version also results in a further performance increase. \textit{Our XL model} dominates our tiny version in each individual dataset.

\section{Conclusion}

In this study, we work on efficient pre-training for face and body detection models in drawings. First of all, we introduce a self-supervised teacher-student network to the domain of drawings. We propose a modified OHEM loss to overcome the false-negative cases caused by the teacher network and equalize the student network's weights to the teacher network's per 500 iterations to prevent distortions in the student network. 

By leveraging the existing style-transferring methods, we highlight the importance of using pre-trained weights for the domain adaptation task and the positive effects of using style-transfer on the pre-training data. Additionally, we analyze the individual impacts of the variations and show that using multiple style-transferring variations together provides higher performance.

Lastly, we train fully supervised models with limited and available labeled data, where the models are initialized with the pre-trained weights. Even with limited drawing data, our model obtains the new SOTA score in most drawing datasets when pre-trained with our pipeline. This finding indicates that efficient pre-training is an important aspect where a low amount of data is available, and the teacher-student network is an effective way of pre-training. 

\bibliographystyle{named}
\bibliography{ijcai23}

\ifshowsupplementary

\newpage
\section{Supplementary Material}

\begin{figure*}[ht]
    \centering
    \includegraphics[width=0.96\textwidth]{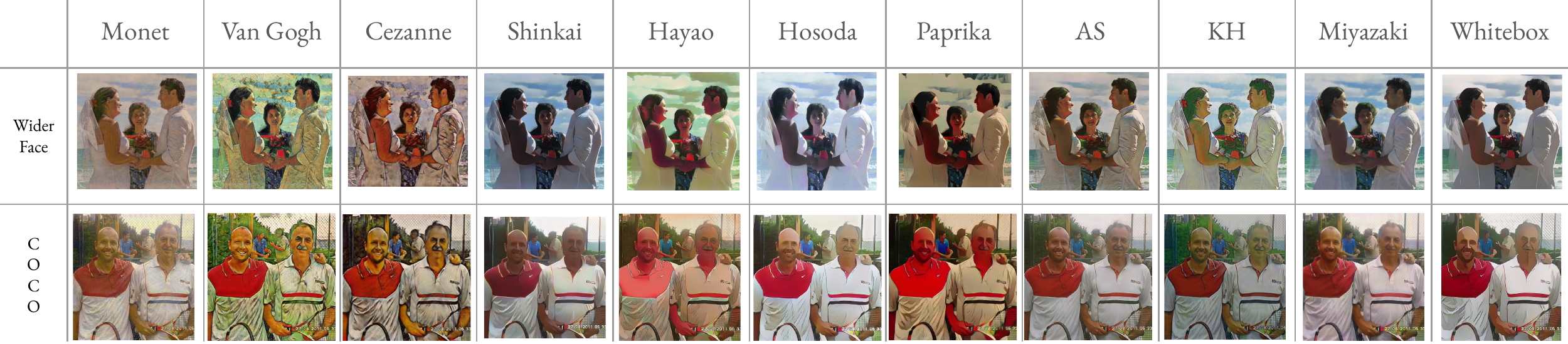}
    \caption{Example results from the Style transfer variations in single WIDER FACE and COCO images.}
    \label{fig:styles_visuals}
\end{figure*}

\input{supplementary_tables/stage_1_results.tex}

In this section, we provide additional explanations on our task, release the extended versions of our experiment tables, and discuss these results. We use abbreviations of datasets in the given tables to save space since there are many datasets for evaluation (iCartoonFace as iCF, Manga 109 faces as M109-F,  Manga 109 bodies as M109-B,  DCM 772 faces as DCM-F, DCM772 bodies as DCM-B, Comic2k as C2k, Watercolor2k as W2k, and Clipart1k as C1k). Average Precision (AP) is selected as the evaluation metric for detection, and the intersection of union value for evaluation is fixed at 0.5. AP scores are calculated by running the same variation five times and computing the average of these runs. At each table given in this section, the best result per column is marked in \textbf{bold} and the second is \underline{underlined} if otherwise is not specified in the table explanation.

\subsection{More on the Applications in Real World}

Drawings are a multi-modal medium for communication. In drawings, stories and thoughts are transmitted mainly through the characters in the scene. Thus, detecting and analyzing faces and bodies are important preliminary tasks for understanding drawings. Through detection, we can process character data to reach the actions, relations, and emotions in the scene, which may lead to reasoning tasks in the next step. By generating a model-aided face and body dataset, we can train generative models for face and character synthesis, face manipulation, real-face-to-drawing-face conversion, etc., which may further help with productization (e.g., animation, comic books, anime, digital art creation) and interfaces for that purpose. A good detector model enables us to perform higher-level tasks in future works. With this motivation, we focus on the detection task and on drawings as our domain in general. However, our design in stage 2 can be adapted to other domain adaptation or generalization studies if a teacher-student-like self-supervised network is utilized. 

\subsection{Results of All Style Transfer Variations}

In the main paper, top-5-scoring style transferring variations are shared. In Table \ref{table:sup_stage_1_results}, you can see performances with the other 6 variations as well. You can also see the visual effect of each style in Figure \ref{fig:styles_visuals}. Although cartoonization methods convert natural images to drawing-like images, they also cause the deterioration of fine details in the images. For the cases where variations result in a worse performance than the non-style-transferred version (i.e., \textit{No Styles}), we observe that these variations are not only inferior in the cartoonization of humans but also cause a decrease in the quality of the image. Thus, utilizing only these variations provides a worse score than no style-transferring.

\subsection{Additional Experiments on Self-Supervised Teacher-Student Network}

In Table \ref{table:sup_stage_2_results}, complete hyper-parameter testing process for stage-2 is given. Since the effects of $\Phi$, $ct^{thres}_{pos}$, $ct^{thres}_{neg}$, $ST$, and $\gamma$ are investigated in the main paper, this section will analyze different values of $d$ (i.e., EMA keep rate), $\beta$ (i.e., coefficient of regression loss), and $c_{teac}$ (i.e., minimum confidence threshold for the student network prediction to be counted as positive in OHEM loss). Moreover, the effect of $\Phi$ in the training curve will also be discussed.

\subsubsection{- EMA Keep Rate ($d$)} 

In the previous self-supervised detection studies, it has been shown that rates below 0.999 result in a worse performance. Therefore, we limit our rate range to $d \in \{0.999, 0.9992, 0.9996, 0.9998, 0.9999\}$. Experiments 2, 9, 10, 11, and 12 contain the model performances with different $d$ values. While we achieve the greatest average performance with $d = 0.9996$, we obtain similar scores with all values except 0.9999.

\subsubsection{- Teacher Confidence Threshold ($c_{teac}$)} 

Stage 2 performance may also significantly change based on different confidence limits for selecting a TN prediction as a pseudo-ground-truth box. Between 2th and 22-25th experiments, we analyze how different teacher confidence thresholds change the AP result. Among the 5 values we set, $c_{teac} = 0.65$ gives the best average AP score among the datasets we evaluate. The outcomes are slightly worse for the values smaller or larger than 0.65. However, the model is not too sensitive to this threshold since the \textit{AP Difference} is only 0.02 for 0.5 and 0.7.

\subsubsection{- Regression Loss Coefficient ($\beta$)}

Between experiments 2 and 13-16, our model's dependence on $\beta$ is investigated. Although the model is not extra sensitive to differing $\beta$ values, the most suitable values for this parameter are 2 and 4.

\begin{figure*}[ht]
    \centering
    \includegraphics[width=0.8\textwidth]{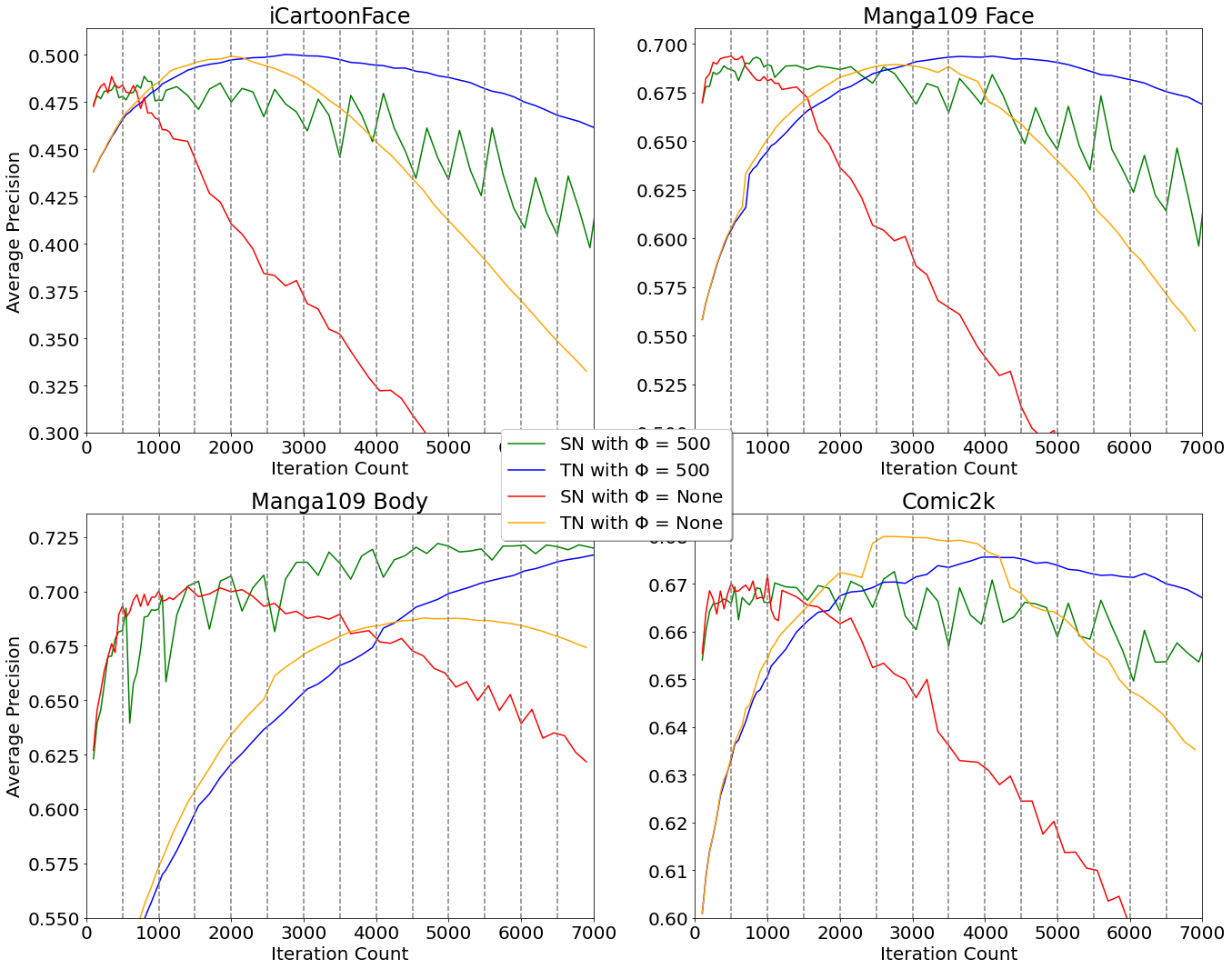}
    \caption{AP curves of teacher and student networks when $\Phi$ is 500 and None.}
    \label{fig:uns_graph}
\end{figure*}

\subsubsection{- Training Curve Analysis ($\Phi$)}

In Figure \ref{fig:uns_graph}, developments of teacher and student networks for 3 different datasets (i.e., iCartoonFace, Manga 109, and Comic 2k) are given. While the y-axis provides the AP scores, the x-axis corresponds to the total training iteration count. As seen in the graphs, in each dataset, the teacher network curve increases more steadily for a longer interval if $\Phi = 500$. Moreover, this setting achieves higher AP scores in iCartoonFace and Manga 109, the two largest and most qualitative labeled drawing datasets. This indicates that setting student network weights with teachers per $\Phi = 500$ iteration results in more stable training and better performance.

\subsection{Stylistic Domain Coverage of Individual Datasets}

In our paper, we express that the labeled drawing data only covers a small subset of the overall domain in terms of stylistic variety. To analyze the stylistic coverage of individual datasets, we design an experiment where the models, that are fine-tuned on a single labeled dataset with a limited number of instances, are evaluated on the other annotated datasets (e.g., if our model is fine-tuned in iCartoonFace, then we also evaluate that model in Manga 109 and DCM 772 faces). By comparing these results, we can infer how valuable each dataset is in other unseen styles/sub-domains. To prevent our model from being affected by other datasets, we will base our statements on the randomly initialized fine-tuning in the Table \ref{table:sup_stage_3_results} (i.e., the columns titled with $N$) during our analysis if no other pre-training is mentioned.

In the following parts, we will use the notation $train\_data_{number\_of\_images}^{eval\_data}$ to mention the models and their results (e.g., if a model is trained in iCartoonFace with 64 images and evaluated on DCM faces, then the notation will be $icf_{64}^{dcm}$).

\input{supplementary_tables/stage_2_results.tex}

\subsubsection{- In Face Data} 

In facial analysis, we observe that our fine-tuned models with iCartoonFace and Manga 109 perform better in other datasets than the variations with DCM 772. In both iCartoonFace and Manga 109, when trained with one dataset (i.e., source sub-domain) and evaluated on the other (i.e., target sub-domain), using 1024 images from the source sub-domain is almost equal to utilizing 128 images from the target sub-domain for training (i.e., $icf_{1024}^{m109} \approx m109_{128}^{m109}$ and $m109_{1024}^{icf} \approx icf_{128}^{icf}$). If all instances are allowed to be leveraged, this equality changes to All \& 512 for both. However, the models trained with DCM 772 obtain significantly worse performances in both iCartoonFace and Manga 109 (e.g., $dcm_{All}^{icf} < icf_{128}^{icf}$ and $dcm_{All}^{m109} \approx m109_{64}^{m109}$). Lastly, Manga 109 models outperform iCartoonFace models on DCM 772, if all of the training data are leveraged (i.e., $m109_{All}^{dcm} > icf_{All}^{dcm}$). In conclusion, while DCM 772 is the worst choice for the unseen sub-domains, both iCartoonFace and Manga 109 result similarly on the other. However, Manga 109 distinguishes further compared to iCartoonFace due to its greater score on DCM 772.


\subsubsection{- In Body Data} 

In the case of no pre-training (N), DCM 772 outperforms C2k* on Manga 109 if the number of images is less than equal to 256. However, with the increasing data size, both dataset models result in similar scores. On the other hand, C2k* outperforms Manga 109 if both dataset models are evaluated on DCM 772 with the number of instances less than equal to 1024. Lastly, Manga 109 outperforms DCM 772 on C2k* evaluation if image instances are between 128 and 1024. Therefore, there is no outstanding winner when no pre-training is applied. Similar results (i.e., no significant difference) are also present for other pre-training options (e.g., $m109_{1024}^{dcm} > c2k*_{1024}^{dcm}$ and $m109_{1024}^{c2k*} < dcm_{1024}^{c2k*}$ and $c2k*_{1024}^{m109} > dcm_{1024}^{m109}$ on SS, $m109_{1024}^{dcm} > c2k*_{1024}^{dcm}$ and $m109_{1024}^{c2k*} < dcm_{1024}^{c2k*}$ and $c2k*_{1024}^{m109} < dcm_{1024}^{m109}$ on ST). Only on style transferred pre-training (ST), DCM 773 dominates others with a small gap.

\subsection{Effect of Pre-Training on Low and High Amount of Data}

In this Section, we examine the effects of our pre-training strategies on stage 3 fine-tuning performance. We execute fine-tuning and evaluation on the same dataset and discuss the influence of both low and high amounts of training data.

While pre-training is always advantageous against random initialization, additional self-supervised pre-training outperforms style-transferred pre-training mainly on the small training data. Additionally, our pre-training design achieves higher scores than previous supervised SOTA models, even with few images. In Table \ref{table:sup_stage_3_results}, we share the model evaluations after fine-tuning with single datasets and a limited number of instances. The scores are \underline{underlined} if they are greater than the previous supervised SOTA drawing detectors.

Self-supervised pre-training before fine-tuning improves detection performance, especially in low amounts of data, but it is still beneficial for higher data sizes compared to random initialization. When trained with only 64 images, there is up to $\sim7\%$ difference between initializing with style-transferred stage 1 weights and self-supervised stage 2 weights. The difference increases up to $\sim35\%$ against random initialization. On the other hand, if the size of the data increases, the margin between different initialization methods decreases. But still, the performance with random initialization on no data limitation is approximately $\sim2\%$ worse than self-supervised stage 2 and style transferred stage 1. This indicates that pre-training is essential for higher performance even with high data.

We obtain better performances than most previous supervised SOTAs even with tiny subsets of the datasets during fine-tuning. 256 panels from Manga 109 are enough to outperform the previous SOTA. This changes to 1024 panels in DCM 772 and 1024 images C2k*. However, we fail to pass the iCartoonFace SOTA. While we aim to keep the model size and design more straightforward, the current iCartoonFace SOTA (ACFD) is $\times5$ times larger than our model and is designed explicitly for the iCartoonFace challenge. Still, our final performance is only $\sim3\%$ smaller. This gap decreases to $0.93\%$ if our model size changes to the XL version of YOLOX.

\input{supplementary_tables/stage_3_results.tex}

\subsection{Visual Results}

In Figures \ref{fig:m109_samples}, \ref{fig:comics_samples}, and \ref{fig:icf_samples} you can see our model's visual outputs for sample drawing images from Manga 109, COMICS, and iCartoonFace. These experiments are conducted with a 0.65 confidence threshold and 0.4 NMS threshold. Overall, moving from stage 1 to stage 2 weights results in a significant increase in detected areas. However, the false positive proposals also increase with this step. Fine-tuning the model that is initialized with the stage 2 weights successfully suppresses these false positive predictions. By increasing the model size, undetected faces, and bodies are further found.

\begin{figure*}
    \centering
    \includegraphics[width=0.82\textwidth]{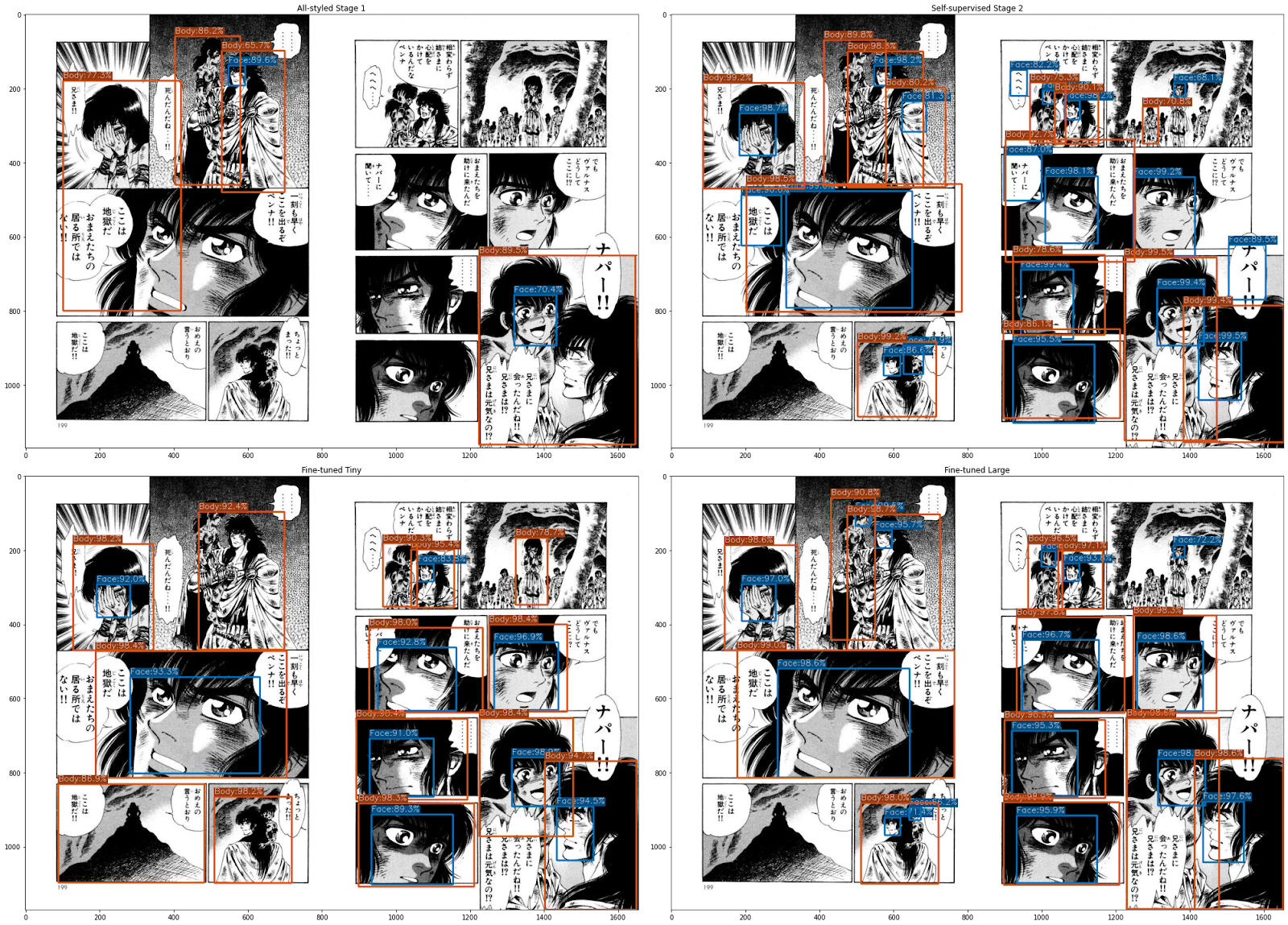}
    \includegraphics[width=0.82\textwidth]{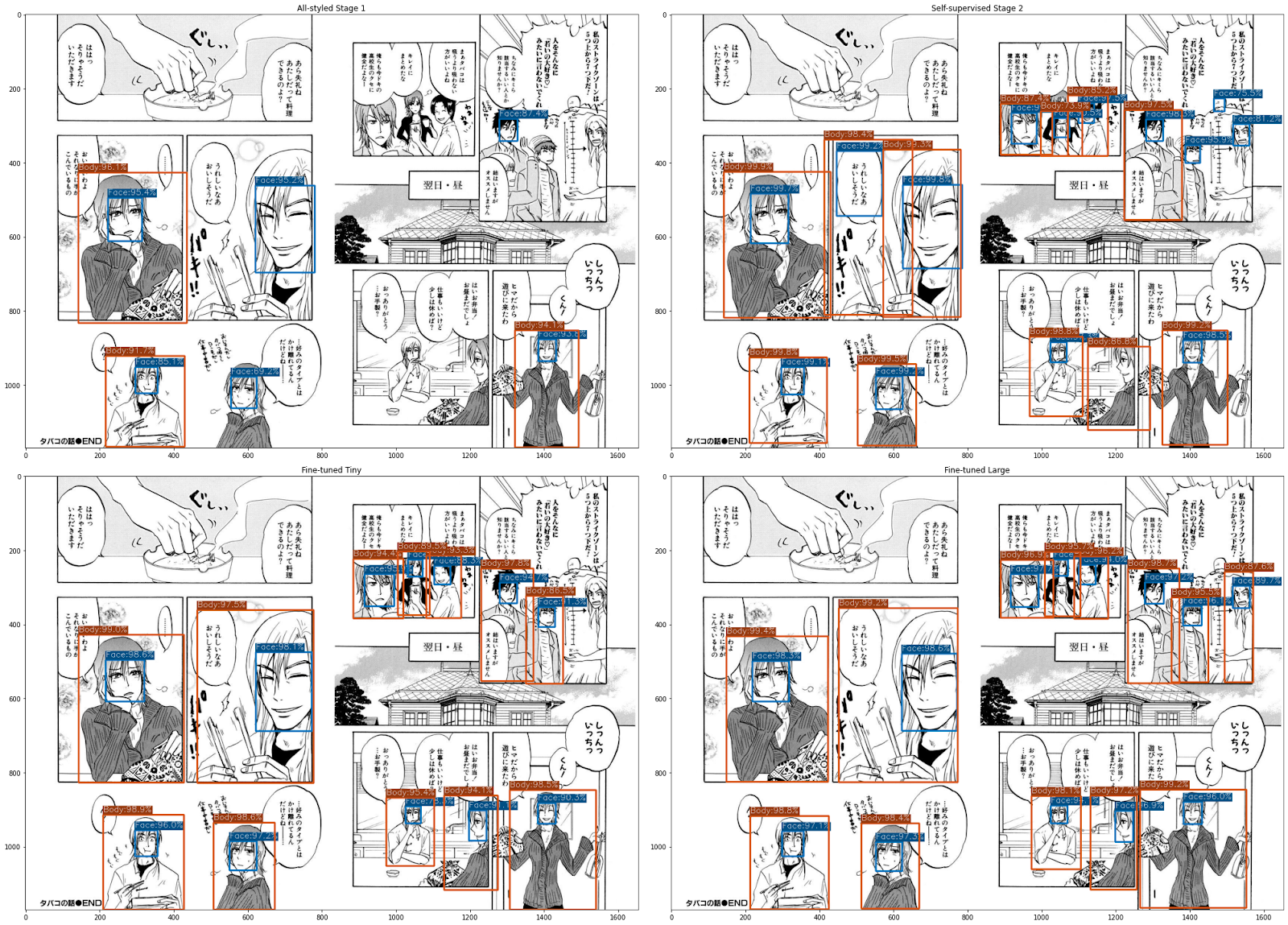}
    \caption{Sample results from Manga 109 pages. Top-left: stage 1 weights, top-right: stage 2 weights, bottom-left: stage 3 weights, bottom-right: stage 3 XL model weights. Better viewed by zooming.}
    \label{fig:m109_samples}
\end{figure*}

\begin{figure*}
    \centering
    \includegraphics[width=0.9\textwidth]{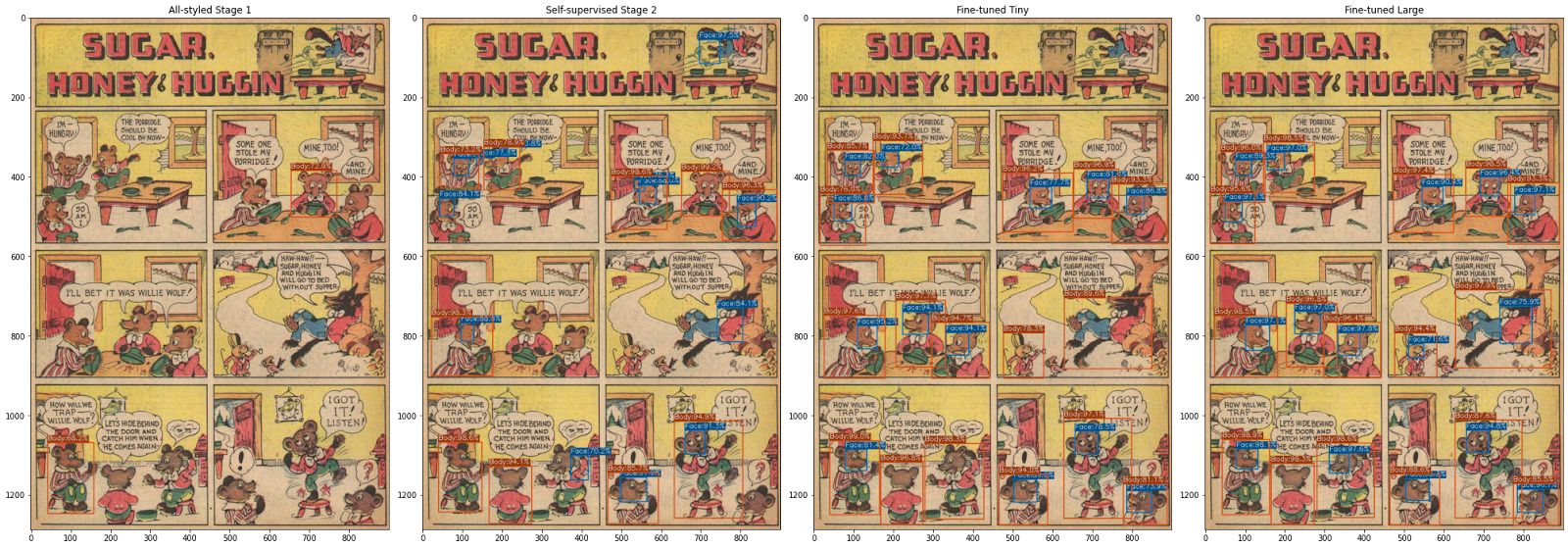}
    \includegraphics[width=0.9\textwidth]{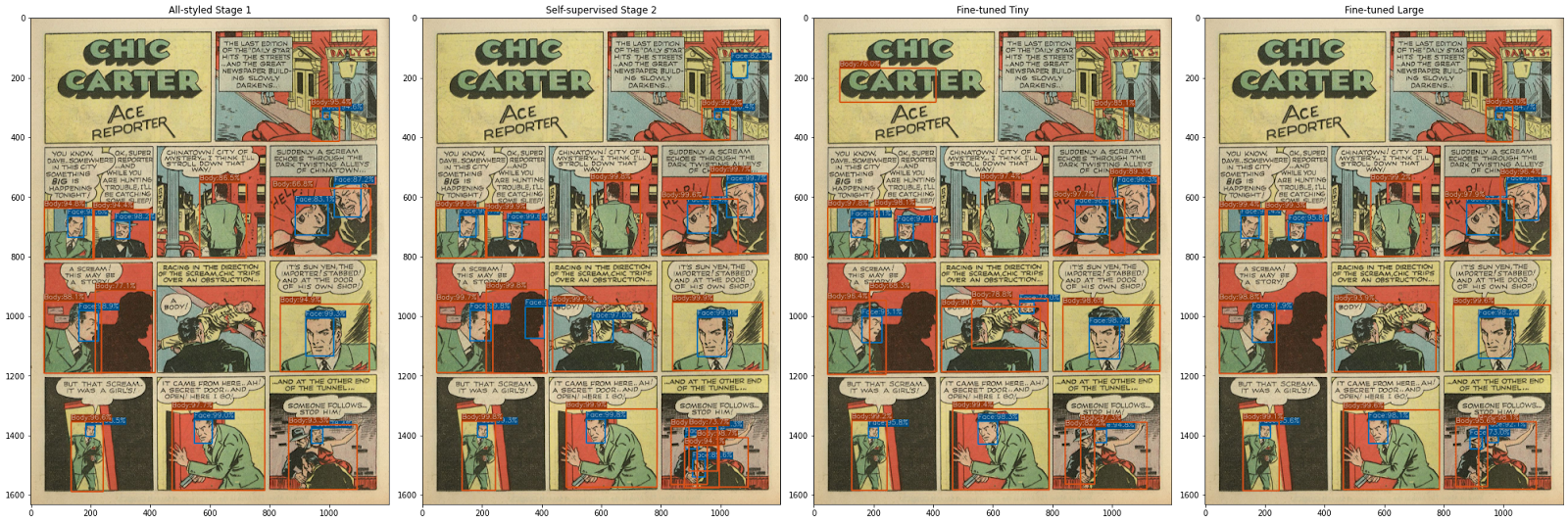}
    \caption{Sample results from COMICS pages. Left to right: stage 1 weights, stage 2 weights, stage 3 weights, stage 3 XL model weights. Better viewed by zooming.}
    \label{fig:comics_samples}
\end{figure*}

\begin{figure*}
    \centering
    \includegraphics[width=0.9\textwidth]{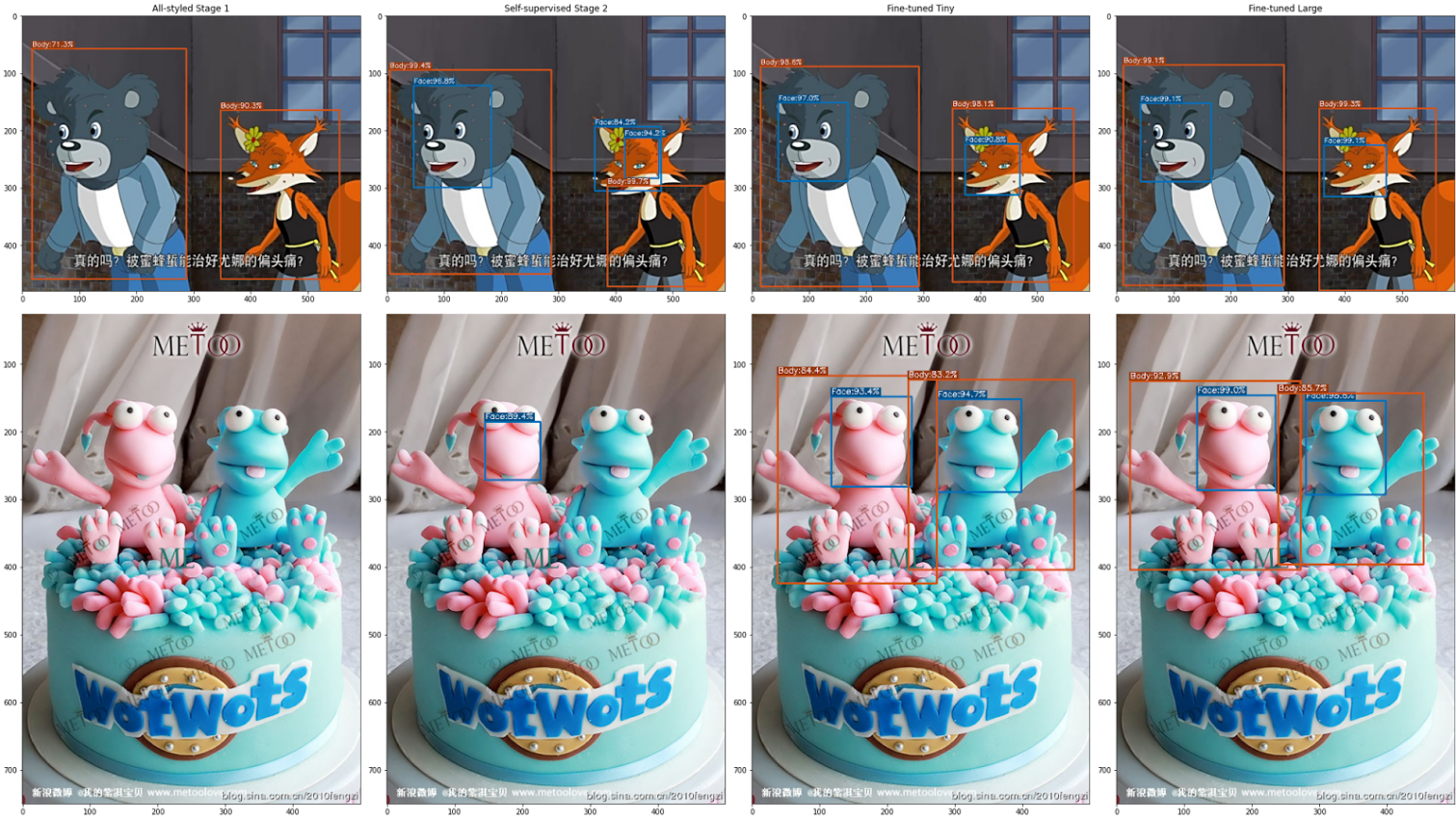}
    \caption{Sample results from iCartoonFace. Left to right: stage 1 weights, stage 2 weights, stage 3 weights, stage 3 XL model weights. Better viewed by zooming.}
    \label{fig:icf_samples}
\end{figure*}

\fi

\end{document}

%% file: tables/stage_2_results.tex
\begin{table*}
\centering
\begin{tabular}{c|cccccc|ccc|c}
    \hline
    \noalign{\smallskip}
    
    \textbf{Index} & \textbf{$\Phi$} & \textbf{Loss} 
    & \textbf{$ct^{thres}_{pos}$} & \textbf{$ct^{thres}_{neg}$} & \textbf{ST} & \textbf{$\gamma$}
    & \textbf{iCF} & \textbf{Manga 109} & \textbf{DCM-B} & \textbf{AP Diff.} \\
    
    \noalign{\smallskip}
    \hline
    \noalign{\smallskip}
    
    1 & 250 & OHEM & 0.15 & 0.85 & Yes & 0.0 & 49.10  & 69.21 & 77.52 & 0.12 \\
    
    2 & 500 & OHEM & 0.15 & 0.85 & Yes & 0.0 & 49.05  & \textbf{69.32} & 77.83 &  0.09 \\
    
    3 & 1000 & OHEM & 0.15 & 0.85 & Yes & 0.0 & 48.48  &	69.02  & \textbf{77.93} & 0.52 \\
    
    4 & Never & OHEM & 0.15 & 0.85 & Yes & 0.0 & 47.83  & 67.68 & 77.29 & 1.56 \\

    \noalign{\smallskip}
    \hline
    \noalign{\smallskip}
    
    5 & 500 & SimOTA & - & - & Yes & 0.0 & 47.13  & 65.64 & 75.42 & 2.89 \\
    
    6 & Never & SimOTA & - & - & Yes & 0.0 & 47.10  & 65.71 & 75.48  &  2.87 \\
    
    \noalign{\smallskip}
    \hline
    \noalign{\smallskip}
    
    7 & 500 & OHEM & 0.70 & 0.30 & Yes & 0.0 & 49.14 & 69.20 & 77.63 & 0.10 \\
    
    8 & 500 & OHEM & 0.50 & 0.50 & Yes & 0.0 & \underline{49.19} & \textbf{69.32} & \underline{77.90} & \textbf{0.02} \\
     
    9 & 500 & OHEM & 0.30 & 0.70 & Yes & 0.0 & 49.09 & \textbf{69.32} & \underline{77.90} & 0.07 \\
    
    10 & 500 & OHEM & 0.00 & 1.00 & Yes & 0.0 & \textbf{49.22} & 69.20 & 77.75 & \underline{0.06} \\

    \noalign{\smallskip}
    \hline
    \noalign{\smallskip}

    11 & 500 & OHEM & 0.15 & 0.85 & No & 0.0
    & 41.66  & 62.64  & 75.64 & 7.12 \\

    \noalign{\smallskip}
    \hline
    \noalign{\smallskip}

    12 & 500 & OHEM & 0.15 & 0.85 & Yes & 0.9
    & 48.72 & 65.72 & 76.59 & 2.05 \\ 

    \noalign{\smallskip}
    \hline
\end{tabular}
\caption{AP scores of different stage 2 configurations in the largest 3 drawing datasets. $\Phi$ is the number of iterations where teacher weights are loaded to student networks afterward, $ct^{thres}_{pos}$ is the minimum confidence threshold for the student network prediction to be counted as positive in ohem loss, $ct^{thres}_{neg}$ is the maximum confidence threshold for the student network prediction to be counted as negative in ohem loss. ST indicates if style transfer is applied in pre-training stage 1, $\gamma$ is the momentum value that is used in the SGD optimizer (if used, nesterov SGD is utilized). The “AP Diff.” column is calculated by averaging the maximum score in each dataset minus the experiment score.}
\label{table:uns_exps_params}
\end{table*}

%% file: tables/stage_1_results.tex
\begin{table}
\centering
\begin{tabular}{l|ccc}
    \hline
    \noalign{\smallskip}
    \textbf{Styles} & \textbf{iCF} & \textbf{Manga 109} 
    & \textbf{DCM-B} \\
    \noalign{\smallskip}
    \hline
    \noalign{\smallskip}
    \textbf{Hayao}    
    & 36.53 {\tiny$\pm$ 0.77} & 35.30 {\tiny$\pm$ 3.66} & 51.63 {\tiny$\pm$ 4.48} \\
    
    \textbf{Shinkai}  
    & 34.88 {\tiny$\pm$ 1.26} & 36.03 {\tiny$\pm$ 2.33} & 56.40 {\tiny$\pm$ 1.85}\\
    
    \textbf{Hosoda}   
    & 38.81 {\tiny$\pm$ 0.40} & 43.05 {\tiny$\pm$ 0.96} & 54.63 {\tiny$\pm$ 3.13} \\
    
    \textbf{KH}       
    & 37.69 {\tiny$\pm$ 0.62} & 36.39 {\tiny$\pm$ 1.43} & 49.10 {\tiny$\pm$ 1.41} \\
    
    \textbf{Whitebox} 
    & 42.22 {\tiny$\pm$ 1.49} & 45.86 {\tiny$\pm$ 1.93} & 52.46 {\tiny$\pm$ 2.23}\\
    
    \noalign{\smallskip}
    \hline
    \noalign{\smallskip}
    
    \textbf{No Styles}  
    & 33.00 {\tiny$\pm$ 1.97} & 35.57 {\tiny$\pm$ 2.82} & 58.94 {\tiny$\pm$ 3.75} \\

    \textbf{Top-5 Styles} 
    & 42.04 {\tiny$\pm$ 1.41} & \underline{47.90} {\tiny$\pm$ 2.61} & 59.96 {\tiny$\pm$ 1.82} \\
    
    \textbf{No Animals} 
    & \underline{42.31} {\tiny$\pm$ 0.70} & 44.81 {\tiny$\pm$ 1.30} & \underline{62.85} {\tiny$\pm$ 1.16} \\

    \textbf{All Styles} 
    & \textbf{42.50} {\tiny$\pm$ 1.25} & \textbf{48.73} {\tiny$\pm$ 2.60} & \textbf{65.46} {\tiny$\pm$ 1.35}  \\
    
    \noalign{\smallskip}
    \hline
\end{tabular}
\caption{AP scores after stage 1 in the largest 3 drawing datasets.}
\label{table:stage1_results}
\end{table}

%% file: tables/stage_3_results.tex
\begin{table*}
\centering
\begin{tabular}{cl|ccc|cccccc}
    \hline
    \noalign{\smallskip}
    \textbf{Types} & \textbf{Models} & 
    \textbf{iCF} & \textbf{M109-F} & \textbf{DCM-F} & 
    \textbf{M109-B} & \textbf{DCM-B} & \textbf{C2k} & \textbf{W2k} & \textbf{C1k} \\
    \noalign{\smallskip}
    \hline
    \noalign{\smallskip}
    NS & \textit{All Styles} 
    & 42.50 & 54.74 & 69.93 & 42.72 & 65.46 & 56.80 & 67.36 & 55.65  \\
    SS & \textit{Teacher-Student} 
    & 49.19 & 69.25 & \textbf{82.45} & 69.38 & 77.90 & 67.41 & 71.53 & 64.25 \\

    \noalign{\smallskip}
    \hline
    \noalign{\smallskip} 
    SS & UMT \cite{umt_model} & - & - & - & - & - & - & 69.90 & 70.50 \\
    SS & D-adapt \cite{dadapt} & - & - & - & - & - & 53.50 & 68.90 & 69.30 \\
    WS & Inoue et al. \cite{inoue2018crossdomain} 
    & - & - & - & 36.71* & 41.89* & 57.30 & 73.20 & 63.00 \\
    WS & H$^2$FA R-CNN \cite{h2farcnn} & - & - & - & - & - & 66.80 & 73.80 & 75.70 \\
    
    \noalign{\smallskip}
    \hline
    \noalign{\smallskip}
    
    FS & \textit{Train w/ 64 Images **} 
    & 65.47 & 80.41 & 69.80  & 77.72 & 77.28  & 68.36 & 71.24 & 58.74  \\

    FS & \textit{Train w/ 256 Images **} 
    &  71.24 & 84.20 & 73.72 & 80.79 & 80.91 & 69.96 & 73.83 & 65.18 \\
    
    FS & \textit{Train w/ 512 Images} **
    & 74.39
    & 85.15
    & 74.85
    & 82.32
    & 82.40
    & 71.05
    & 77.63
    & - \\
    
    FS & \textit{Train w/ All Images **}
    & 87.75
    & \underline{87.86}
    & 75.87
    & \underline{87.06}
    & \underline{84.89}
    & \underline{71.66}
    & \underline{89.17}
    & \underline{77.97}\\

    \noalign{\smallskip}
    \hline
    \noalign{\smallskip}

    FS & \textit{XL Model w/ All Images **} 
    & \underline{90.01}
    & \textbf{87.88}
    & \underline{77.40}
    & \textbf{87.98}
    & \textbf{86.14}
    & \textbf{73.65}
    & \textbf{89.81}
    & \textbf{83.59} \\ 
    
    \noalign{\smallskip}
    \hline
    \noalign{\smallskip}
    FS & ACFD \cite{zhang2020acfd} & \textbf{90.94} & - & - & - & - & - & - & - \\
    FS & Ogawa et al. \cite{ogawa2018object} & - & 76.20 & - & 79.60 & - & - & - & - \\
    FS & Nguyen et al. \cite{jimaging4070089} & - & - & 74.94 & - & 76.76  & - & - & - \\
    FS & Inoue et al. \cite{inoue2018crossdomain} & - & - & - & - & - & 70.10 & 77.30 & 76.20 \\
    \noalign{\smallskip}
    \hline
\end{tabular}
\caption{Overall AP performances of our models and previous SOTA models. Our models are titled in \textit{italic}. The teacher-student network is initialized with the style transferred pre-training, All of our supervised models are initialized with pre-training stage 2 weights. NS: no target domain supervision. SS: self-supervision, WS: weak-supervision, FS: full target domain supervision. Scores with "*" mean that they are evaluated by us using the model from the original project repository. "**" indicates that the results are retrieved from single-dataset trainings and each score is calculated by a separate model trained specifically with the particular dataset.}
\label{table:models_perf}
\end{table*}

%% file: tables/pretrain_comparisons.tex
\begin{table}
\centering
\begin{tabular}{l|ccccc|c}
    \hline
    \noalign{\smallskip}
    
    & \multicolumn{3}{c}{\textbf{Image Instance Counts}} \\
    
    \textbf{Pre-training} 
    & \textbf{64} 
    & \textbf{512} 
    & \textbf{All} \\
    
    \noalign{\smallskip}
    \hline
    \noalign{\smallskip}
    
    \textbf{None}  
    & 47.79 {\tiny$\pm$ 1.38}  
    & 69.38 {\tiny$\pm$ 0.82}
    & 80.60 {\tiny$\pm$ 0.65} \\
    
    \textbf{Stage 1} 
    & 66.90 {\tiny$\pm$ 1.40}
    & 75.34 {\tiny$\pm$ 0.99}
    & \textbf{82.87} {\tiny$\pm$ 1.53} \\
    
    \textbf{Stage 1 + 2} 
    & \textbf{71.13} {\tiny$\pm$ 0.92}
    & \textbf{77.44} {\tiny$\pm$ 0.47}
    & 82.78 {\tiny$\pm$ 0.93} \\
    
    \noalign{\smallskip}
    \hline
\end{tabular}
\caption{Average AP performance of our model when trained with a subset of individual datasets having annotations of a limited number of random images. Average is calculated by taking the mean of each score retrieved from each 6 datasets.}
\label{table:single_best_avg}
\end{table}

%% file: supplementary_tables/stage_1_results.tex
\setlength{\tabcolsep}{4pt}
\begin{table*}
\centering
\begin{tabular}{l|ccc|ccccc}
    \hline
    \noalign{\smallskip}
    \textbf{Styles} & \textbf{iCF} & \textbf{M109-F} & \textbf{DCM-F} & \textbf{M109-B} & \textbf{DCM-B} & \textbf{C2k} & \textbf{W2k} & \textbf{C1k} \\
    \noalign{\smallskip}
    \hline
    \noalign{\smallskip}
    \textbf{Hayao}    & 36.53 {\tiny$\pm$ 0.77}&42.34 {\tiny$\pm$ 3.44}&63.43 {\tiny$\pm$ 1.44}&28.25 {\tiny$\pm$ 3.88}&51.63 {\tiny$\pm$ 4.48}&47.88 {\tiny$\pm$ 1.11}&61.38 {\tiny$\pm$0.97}&49.07 {\tiny$\pm$1.28} \\
    \textbf{Shinkai}  & 34.88 {\tiny$\pm$ 1.26}&41.21 {\tiny$\pm$ 2.64}&57.56 {\tiny$\pm$ 3.11}&30.84 {\tiny$\pm$ 2.01}& {56.40} {\tiny$\pm$ 1.85}&48.21 {\tiny$\pm$ 1.15}&60.26 {\tiny$\pm$1.21}&50.47 {\tiny$\pm$0.65} \\
    \textbf{Hosoda}   & 38.81 {\tiny$\pm$ 0.40}&49.59 {\tiny$\pm$ 0.85}&60.35 {\tiny$\pm$ 3.13}&36.50 {\tiny$\pm$ 1.07}&54.63 {\tiny$\pm$ 3.13}&51.12 {\tiny$\pm$ 0.90}&62.52 {\tiny$\pm$0.67}&53.58 {\tiny$\pm$1.22} \\
    \textbf{Paprika}  & 32.27 {\tiny$\pm$ 0.66}&32.16 {\tiny$\pm$ 4.17}&50.95 {\tiny$\pm$ 3.39}&21.69 {\tiny$\pm$ 1.84}&40.40 {\tiny$\pm$ 3.09}&40.50 {\tiny$\pm$ 1.64}&47.96 {\tiny$\pm$2.43}&40.46 {\tiny$\pm$1.58} \\
    \textbf{Van Gogh} & 33.31 {\tiny$\pm$ 1.66}&35.30 {\tiny$\pm$ 1.80}&{62.73} {\tiny$\pm$ 1.30}&26.09 {\tiny$\pm$ 2.42}&50.80 {\tiny$\pm$ 2.56}&44.66 {\tiny$\pm$ 1.87}&58.38 {\tiny$\pm$1.16}&46.18 {\tiny$\pm$1.68} \\
    \textbf{Monet}    & 26.25 {\tiny$\pm$ 2.98}&30.44 {\tiny$\pm$ 3.06}&58.89 {\tiny$\pm$ 4.06}&21.13 {\tiny$\pm$ 2.98}&50.96 {\tiny$\pm$ 1.54}&39.31 {\tiny$\pm$ 1.81}&58.84 {\tiny$\pm$1.43}&44.62 {\tiny$\pm$2.13} \\
    \textbf{Cezanne}  & 29.96 {\tiny$\pm$ 0.76}&35.49 {\tiny$\pm$ 2.64}&59.10 {\tiny$\pm$ 2.59}&26.76 {\tiny$\pm$ 3.05}&46.22 {\tiny$\pm$ 8.25}&41.50 {\tiny$\pm$ 4.16}&52.04 {\tiny$\pm$8.21}&42.12 {\tiny$\pm$4.08} \\
    \textbf{Miyazaki} & 32.16 {\tiny$\pm$ 1.62}&38.39 {\tiny$\pm$ 0.94}&59.63 {\tiny$\pm$ 2.52}&28.31 {\tiny$\pm$ 1.53}&55.91 {\tiny$\pm$ 2.23}&42.78 {\tiny$\pm$ 0.29}&61.31 {\tiny$\pm$0.45}&47.83 {\tiny$\pm$1.71} \\
    \textbf{AS}       & 35.34 {\tiny$\pm$ 0.94}&39.81 {\tiny$\pm$ 2.83}&57.09 {\tiny$\pm$ 0.66}&27.01 {\tiny$\pm$ 1.34}&52.23 {\tiny$\pm$ 2.19}&44.79 {\tiny$\pm$ 1.81}&60.81 {\tiny$\pm$0.64}&47.34 {\tiny$\pm$0.69} \\
    \textbf{KH}       & 37.69 {\tiny$\pm$ 0.62}&44.18 {\tiny$\pm$ 2.20}& 59.68 {\tiny$\pm$ 2.97}&28.59 {\tiny$\pm$ 0.65}&49.10 {\tiny$\pm$ 1.41}&49.04 {\tiny$\pm$ 1.14}& {63.51} {\tiny$\pm$1.00}&50.34 {\tiny$\pm$0.73} \\
    \textbf{Whitebox} & {{42.22}} {\tiny$\pm$ 1.49}& {{53.10}} {\tiny$\pm$ 2.12}&59.63 {\tiny$\pm$ 3.93}& {{38.61}} {\tiny$\pm$ 1.73}&52.46 {\tiny$\pm$ 2.23}& {52.41} {\tiny$\pm$ 1.25}&63.35 {\tiny$\pm$0.41}& {54.56} {\tiny$\pm$1.64} \\
    \noalign{\smallskip}
    \hline
    \noalign{\smallskip}
    \textbf{No Styles}  &33.00 {\tiny$\pm$ 1.97}&40.44 {\tiny$\pm$ 3.04}&59.63 {\tiny$\pm$ 3.83}&30.69 {\tiny$\pm$ 2.59}&58.94 {\tiny$\pm$ 3.75}& 44.81 {\tiny$\pm$ 1.05}& 61.88 {\tiny$\pm$ 0.95}& 49.60 {\tiny$\pm$ 1.20} \\
    \textbf{Top-5 Styles}
    & 42.04 {\tiny$\pm$ 1.41} & \underline{53.82} {\tiny$\pm$ 3.28} & {65.94} {\tiny$\pm$ 2.71}
    & \underline{41.98} {\tiny$\pm$ 1.94} & {59.96} {\tiny$\pm$ 1.82} & \underline{55.36} {\tiny$\pm$ 1.56}
    & {66.91} {\tiny$\pm$ 0.89} & \underline{55.99} {\tiny$\pm$ 0.73} \\
    \textbf{No Animals} &\underline{42.31} {\tiny$\pm$ 0.70}&52.09 {\tiny$\pm$ 1.73}&\underline{69.70} {\tiny$\pm$ 2.26}&37.53 {\tiny$\pm$ 0.87}&\underline{62.85} {\tiny$\pm$ 1.16}& {54.58} {\tiny$\pm$ 0.51}& \textbf{67.97} {\tiny$\pm$ 0.24}& \textbf{58.37} {\tiny$\pm$ 0.64} \\
    \textbf{All Styles} &\textbf{42.50} {\tiny$\pm$ 1.25}&\textbf{54.74} {\tiny$\pm$ 2.20}&\textbf{69.93} {\tiny$\pm$ 2.67}&\textbf{42.72} {\tiny$\pm$ 3.00}&\textbf{65.46} {\tiny$\pm$ 1.35}& \textbf{56.80} {\tiny$\pm$ 1.42}& \underline{67.36} {\tiny$\pm$ 0.39}& {55.65} {\tiny$\pm$ 0.87} \\
    \noalign{\smallskip}
    \hline
\end{tabular}
\caption{AP performances of our model after pre-training stage 1 for different style transferring variations: with single style transferring variation is selected, no style is selected, top-5 best-performing styles are combined, all styles are combined but no animal annotations are included, all styles are combined and animal annotations are also included.}
\label{table:sup_stage_1_results}
\end{table*}

%% file: supplementary_tables/stage_2_results.tex
\begin{table*}
\centering
\fontsize{8.5}{12}\selectfont
\Rotatebox{270}{%
\begin{tabular}{c|ccccccccc|ccc|ccccc|c}
    \hline
    \noalign{\smallskip}
    
    \textbf{Index} & \textbf{$\Phi$} & \textbf{Loss} & \textbf{$d$} & \textbf{$\beta$} & \textbf{$ct^{thres}_{pos}$} & \textbf{$ct^{thres}_{neg}$} & \textbf{$c_{teac}$}
    & \textbf{ST} & \textbf{$\gamma$}& \textbf{iCF} & \textbf{M109-F} & \textbf{DCM-F} 
    & \textbf{M109-B} & \textbf{DCM-B} & \textbf{C2k} & \textbf{W2k} & \textbf{C1k} & \textbf{AP Diff.}\\
    \noalign{\smallskip}
    \hline
    \noalign{\smallskip}
    
    1 & 250 & OHEM & 0.9996 & 2 & 0.15 & 0.85 & 0.65 & Yes & 0.0
    & 49.10  & 69.14  & 81.52 
    & 69.28  & 77.52  & 67.13 
    & 71.18  & {64.63}  & 0.697 \\
    
    2 & 500 & OHEM & 0.9996 & 2 & 0.15 & 0.85 & 0.65 & Yes & 0.0
    & 49.05  & 69.23  &  82.22
    & 69.41  & 77.83  & 67.38
    & 71.60  & 64.12  & 0.433 \\
    
    3 & 1000 & OHEM & 0.9996 & 2 & 0.15 & 0.85 & 0.65 & Yes & 0.0
    & 48.48  &	68.92  & 82.14	 
    & 69.11  &	\underline{77.93}  & 67.30  
    & 72.18  &	63.37  & 0.527 \\
    
    4 & 2000 & OHEM & 0.9996 & 2 & 0.15 & 0.85 & 0.65 & Yes & 0.0
    & 48.71  & 68.79  & 81.91  
    & 67.30  & 77.50  & 66.79  
    & 72.07  & 63.83  & 0.954 \\	
   
    5 & 5000 & OHEM & 0.9996 & 2 & 0.15 & 0.85 & 0.65 & Yes & 0.0
    & 48.26  & 68.55  & 81.62  
    & 67.04  & 77.30  & 66.88 
    & 72.37  & 63.23  & 1.104 \\
    
    6 & Never & OHEM & 0.9996 & 2 & 0.15 & 0.85 & 0.65 & Yes & 0.0
    & 47.83  & 68.32  & 81.71 
    & 67.03  & 77.29  & 67.21 
    & 72.33  & 63.27  & 1.147 \\

    \noalign{\smallskip}
    \hline
    \noalign{\smallskip}
    
    7 & 500 & SimOTA & 0.9996 & 2 & - & - & 0.65 & Yes & 0.0
    & 47.13  & 67.65  &  82.23
    & 63.62  & 75.42  & 65.86 
    & \underline{72.55}  & 61.41  & 2.184 \\
    
    8 & Never & SimOTA & 0.9996 & 2 & - & - & 0.65 & Yes & 0.0
    & 47.10  & 67.58  & 82.19 
    & 63.83  & 75.48  & 65.96  
    & \textbf{72.59}  & 61.36  & 2.146 \\

    \noalign{\smallskip}
    \hline
    \noalign{\smallskip}
    
    9 & 500 & OHEM & 0.9990 & 2 & 0.15 & 0.85 & 0.65 & Yes & 0.0
    & 49.01  & 69.10  & 82.21  
    & \underline{69.48}  & 77.89  & 67.24  
    & 71.30  & 64.21  & 0.503 \\
    
    10 & 500 & OHEM & 0.9992 & 2 & 0.15 & 0.85 & 0.65 & Yes & 0.0
    & 49.05  & 69.07  & 81.95  
    & 69.29  & 77.83  & 67.25  
    & 71.46  & 64.08  & 0.550 \\
    
    11 & 500 & OHEM & 0.9998 & 2 & 0.15 & 0.85 & 0.65 & Yes & 0.0
    & \underline{49.31}  & \underline{69.32}  & 81.74  
    & 68.37  & 77.44  & 67.23  
    & 71.83  & 64.40  & 0.644 \\
    
    12 & 500 & OHEM & 0.9999 & 2 & 0.15 & 0.85 & 0.65 & Yes & 0.0
    & \textbf{49.81}  & 68.31  & 79.28  
    & 63.52  & 75.35  & 65.67  
    & 72.17  & 64.37  & 2.234 \\
    
    \noalign{\smallskip}
    \hline
    \noalign{\smallskip}
    
    13 & 500 & OHEM & 0.9996 & 0 & 0.15 & 0.85 & 0.65 & Yes & 0.0
    & 48.07  & 68.71  & 81.65  
    & 69.13  & 77.60  & 67.02 
    & 71.41  & \underline{64.79}  & 0.880 \\
    
    14 & 500 & OHEM & 0.9996 & 1 & 0.15 & 0.85 & 0.65 & Yes & 0.0
    & {49.29}  & \textbf{69.36}  & 81.64  
    & 68.65  & 77.50  & 66.97 
    & 71.75  & \textbf{65.02}  & 0.656 \\
    
    15 & 500 & OHEM & 0.9996 & 4 & 0.15 & 0.85 & 0.65 & Yes & 0.0
    & 48.82  & 69.04  &  82.44
    & \textbf{69.65}  & 77.82  & \textbf{67.55} 
    & 71.69  & 63.50  & \underline{0.391} \\
    
    16 & 500 & OHEM & 0.9996 & 10 & 0.15 & 0.85 & 0.65 & Yes & 0.0
    & 48.94  & 69.07  & 81.71  
    & 68.43  & 77.87  & 67.31
    & {72.43}  & 63.10  & 0.570 \\
    
    \noalign{\smallskip}
    \hline
    \noalign{\smallskip}
    
    17 & 500 & OHEM & 0.9996 & 2 & 0.70 & 0.30 & 0.65 & Yes & 0.0
    & 49.14 & 69.26 & 82.13 & 69.14 & 77.63 & {67.41} & 71.67 & 64.07 & 0.481 \\
    
    18 & 500 & OHEM & 0.9996 & 2 & 0.50 & 0.50 & 0.65 & Yes & 0.0
    & 49.19 & 69.25 & {82.45} & 69.38 & 77.90 & {67.41} & 71.53 & 64.25 & \textbf{0.377} \\
     
    19 & 500 & OHEM & 0.9996 & 2 & 0.30 & 0.70 & 0.65 & Yes & 0.0
    & 49.09 & 69.15 & 82.08 & \underline{69.48} & 77.90 & 67.37 & 71.68 & 64.09 & 0.429 \\

    20 & 500 & OHEM & 0.9996 & 2 & 0.05 & 0.95 & 0.65 & Yes & 0.0
    & 48.99 & 69.14 & 82.33 & 69.32 & \textbf{78.03} & 67.31 & 71.62 & 63.99 & 0.430 \\
    
    21 & 500 & OHEM & 0.9996 & 2 & 0.00 & 1.00 & 0.65 & Yes & 0.0
    & 49.22 & {69.27} & 82.12 & 69.12 & 77.75 & 67.40 & 71.72 & 64.33 & 0.450 \\
    
    \noalign{\smallskip}
    \hline
    \noalign{\smallskip}
     
    22 & 500 & OHEM & 0.9996 & 2 & 0.15 & 0.85 & 0.35 & Yes & 0.0
    & 49.09 & 69.04 & 81.70 & 69.25 & 77.63 & 67.27 & 71.83 & 64.10 & 0.563 \\
    
    23 & 500 & OHEM & 0.9996 & 2 & 0.15 & 0.85 & 0.50 & Yes & 0.0
    & 49.15 & 69.24 & 81.50 & 69.19 & 77.75 & \underline{67.45} & 71.82 & 64.55 & 0.521 \\
    
    24 & 500 & OHEM & 0.9996 & 2 & 0.15 & 0.85 & 0.75 & Yes & 0.0
    & 49.08 & 69.26 & \underline{82.51} & 69.19 & 77.85 & 67.26 & 71.43 & 64.13 & 0.453 \\
    
    25 & 500 & OHEM & 0.9996 & 2 & 0.15 & 0.85 & 0.90 & Yes & 0.0
    & 48.92 & 69.20 & \textbf{82.76} & 68.95 & \underline{77.93} & 66.93 & 71.04 & 63.67 & 0.575 \\

    \noalign{\smallskip}
    \hline
    \noalign{\smallskip}

    26 & 500 & OHEM & 0.9996 & 2 & 0.15 & 0.85 & 0.65 & Yes & 0.9 
    & 48.72 & 67.26 & 80.10 & 64.17 & 76.59 & 66.81 & 72.51 & 62.20 & 1.941 \\ 

    \noalign{\smallskip}
    \hline
    \noalign{\smallskip}

    27 & 500 & OHEM & 0.9996 & 2 & 0.15 & 0.85 & 0.65 & No & 0.0
    & 41.66  & 58.96 & 74.56 & 66.32 & 75.64 & 65.32 & 69.56 & 64.13 & 5.390 \\

    \noalign{\smallskip}
    \hline
\end{tabular}
}
\caption{AP scores of different unsupervised experiment configurations. $\Phi$: number of iterations where teacher weights are loaded to student networks afterward, $d$: ema keep rate, $\beta$: coefficient of regression loss, $c_{teac}$: confidence threshold of teacher network to select a prediction as pseudo ground truth, $ct^{thres}_{pos}$:  minimum confidence threshold for the student network prediction to be counted as positive in OHEM loss, $ct^{thres}_{neg}$: maximum confidence threshold for the student network prediction to be counted as negative in OHEM loss, $ST$: if style transfer is applied in stage 1 pre-training, $\gamma$: if momentum is used in the optimizer. the “AP Diff.” column is calculated by averaging the maximum score in each dataset minus the experiment score.}
\label{table:sup_stage_2_results}
\end{table*}

%% file: supplementary_tables/stage_3_results.tex
\begin{table*}
\centering
\fontsize{8.5}{12}\selectfont
\begin{tabular}{l|c|ccc|ccc|ccc}
    \noalign{\smallskip}
    \cline{3-11}
    \noalign{\smallskip}
    \multicolumn{1}{c}{} & \multicolumn{1}{c}{} & \multicolumn{9}{c}{\textbf{Evaluation Datasets}} \\
    
    \noalign{\smallskip}
    \hline
    \noalign{\smallskip}
    
    \textbf{Training} & \textbf{\# of} & \multicolumn{3}{c|}{\textbf{iCF}} 
    & \multicolumn{3}{c|}{\textbf{M109-F}} 
    & \multicolumn{3}{c}{\textbf{DCM-F}}  \\
    \textbf{Datasets} & \textbf{images} 
    & \textbf{N} & \textbf{ST} & \textbf{SS}
    & \textbf{N} & \textbf{ST} & \textbf{SS}
    & \textbf{N} & \textbf{ST} & \textbf{SS} \\
    \noalign{\smallskip}
    \hline
    \noalign{\smallskip}

    \textbf{iCF} & \textbf{64}   
    & 42.29 {\tiny$\pm$ 5.18} 
    & 61.67 {\tiny$\pm$ 1.27} 
    & {65.47} {\tiny$\pm$ 0.67} 
    & 36.40 {\tiny$\pm$ 7.84} 
    & 66.50 {\tiny$\pm$ 3.68} 
    & {73.55} {\tiny$\pm$ 1.51} 
    & 26.97 {\tiny$\pm$ 5.77} 
    & 52.40 {\tiny$\pm$ 11.38}
    & {73.31} {\tiny$\pm$ 3.35}  \\
    
    \textbf{iCF} & \textbf{128}
    & 52.18 {\tiny$\pm$ 1.89}
    & 64.41 {\tiny$\pm$ 1.05} 
    & {68.58} {\tiny$\pm$ 0.62} 
    & 48.78 {\tiny$\pm$ 5.57} 
    & 69.00 {\tiny$\pm$ 3.91} 
    & {75.85} {\tiny$\pm$ 1.06} 
    & 31.13 {\tiny$\pm$ 9.91} 
    & 59.36 {\tiny$\pm$ 7.81} 
    & {72.35} {\tiny$\pm$ 4.07} \\
    
    \textbf{iCF} & \textbf{256}   
    & 60.87 {\tiny$\pm$ 0.68} 
    & 69.20 {\tiny$\pm$ 0.91} 
    & {71.24} {\tiny$\pm$ 0.65} 
    & 62.25 {\tiny$\pm$ 2.11} 
    & 74.51 {\tiny$\pm$ 1.88} 
    & {\underline{76.08}} {\tiny$\pm$ 0.67} 
    & 44.11 {\tiny$\pm$ 4.78} 
    & 60.94 {\tiny$\pm$ 8.46} 
    & {70.59} {\tiny$\pm$ 1.85} \\
    
    \textbf{iCF} & \textbf{512}   
    & 66.22 {\tiny$\pm$ 0.49}
    & 72.51 {\tiny$\pm$ 1.38} 
    & {74.39} {\tiny$\pm$ 0.60} 
    & 68.97 {\tiny$\pm$ 3.51} 
    & 75.38 {\tiny$\pm$ 2.59} 
    & {\underline{78.43}} {\tiny$\pm$ 1.46} 
    & 50.91 {\tiny$\pm$ 1.92} 
    & 58.10 {\tiny$\pm$ 9.41} 
    & {66.30} {\tiny$\pm$ 5.25} \\
    
    \textbf{iCF} & \textbf{1024}   
    & 72.47 {\tiny$\pm$ 0.78} 
    & 77.22 {\tiny$\pm$ 2.13} 
    & {77.31} {\tiny$\pm$ 0.30} 
    & 73.36 {\tiny$\pm$ 2.74} 
    & \underline{78.36} {\tiny$\pm$ 3.65} 
    & {\underline{80.59}} {\tiny$\pm$ 0.47} 
    & 53.44 {\tiny$\pm$ 3.90} 
    & 61.43 {\tiny$\pm$ 3.84} 
    & {67.90} {\tiny$\pm$ 2.87} \\
    
    \textbf{iCF} & \textbf{All}
    & 83.70 {\tiny$\pm$ 0.21} 
    & 87.61 {\tiny$\pm$ 0.07} 
    & {\textbf{87.75}} {\tiny$\pm$ 0.02}	
    & \underline{83.33} {\tiny$\pm$ 0.49} 
    & \underline{85.62} {\tiny$\pm$ 0.06} 
    & {\underline{85.63}} {\tiny$\pm$ 0.13}
    & 64.16 {\tiny$\pm$ 0.98} 
    & 71.98 {\tiny$\pm$ 1.16} 
    & {72.11} {\tiny$\pm$ 0.41} \\

    \noalign{\smallskip}
    \hline
    \noalign{\smallskip}
    
    \textbf{M109} & \textbf{64}   
    & 25.28 {\tiny$\pm$ 4.19} 
    & 47.36 {\tiny$\pm$ 4.06} 
    & {51.99} {\tiny$\pm$ 3.04} 
    & 67.46 {\tiny$\pm$ 3.46} 
    & \underline{77.70} {\tiny$\pm$ 4.05} 
    & {\underline{80.41}} {\tiny$\pm$ 1.41} 
    & 22.60 {\tiny$\pm$ 8.16} 
    & 54.40 {\tiny$\pm$ 7.13} 
    & {70.90} {\tiny$\pm$ 3.70} \\
    
    \textbf{M109} & \textbf{128}
    & 34.54 {\tiny$\pm$ 3.29} 
    & 49.40 {\tiny$\pm$ 1.20} 
    & {53.47} {\tiny$\pm$ 1.18} 
    & 74.89 {\tiny$\pm$ 0.82} 
    & \underline{80.35} {\tiny$\pm$ 0.54} 
    & {\underline{82.14}} {\tiny$\pm$ 1.12} 
    & 35.45 {\tiny$\pm$ 5.18} 
    & 56.63 {\tiny$\pm$ 5.65} 
    & {71.02} {\tiny$\pm$ 1.44} \\
    
    \textbf{M109} & \textbf{256}   
    & 39.59 {\tiny$\pm$ 3.96} 
    & 53.11 {\tiny$\pm$ 0.83} 
    & {56.11} {\tiny$\pm$ 1.97} 
    & \underline{76.80} {\tiny$\pm$ 0.49} 
    & \underline{83.35} {\tiny$\pm$ 0.62} 
    & {\underline{84.20}} {\tiny$\pm$ 0.68} 
    & 41.42 {\tiny$\pm$ 10.71}
    & 57.83 {\tiny$\pm$ 3.10} 
    & {71.71} {\tiny$\pm$ 3.12} \\
    
    \textbf{M109} & \textbf{512}   
    & 48.55 {\tiny$\pm$ 1.68} 
    & 58.16 {\tiny$\pm$ 2.48} 
    & {58.74} {\tiny$\pm$ 1.42} 
    & \underline{82.98} {\tiny$\pm$ 0.54} 
    & {\underline{85.58}} {\tiny$\pm$ 0.61} 
    & \underline{85.15} {\tiny$\pm$ 0.22} 
    & 53.83 {\tiny$\pm$ 4.76} 
    & 63.56 {\tiny$\pm$ 4.02} 
    & {74.17} {\tiny$\pm$ 1.86} \\
    
    \textbf{M109} & \textbf{1024}   
    & 51.41 {\tiny$\pm$ 2.40} 
    & 60.80 {\tiny$\pm$ 0.95} 
    & {62.46} {\tiny$\pm$ 0.80} 
    & \underline{85.83} {\tiny$\pm$ 0.41} 
    & {\underline{86.50}} {\tiny$\pm$ 0.27} 
    & \underline{86.21} {\tiny$\pm$ 0.19} 
    & 57.74 {\tiny$\pm$ 4.22} 
    & 67.99 {\tiny$\pm$ 2.21} 
    & {72.87} {\tiny$\pm$ 1.43} \\

    \textbf{M109} & \textbf{All}
    & 66.84 {\tiny$\pm$ 1.56} 
    & 69.89 {\tiny$\pm$ 0.65} 
    & {70.71} {\tiny$\pm$ 0.49}	
    & \underline{87.70} {\tiny$\pm$ 0.05} 
    & {\textbf{87.87}} {\tiny$\pm$ 0.07} 
    & \underline{87.86} {\tiny$\pm$ 0.02}
    & \underline{77.20} {\tiny$\pm$ 1.66} 
    & \underline{75.15} {\tiny$\pm$ 1.36} 
    & {\underline{78.40}} {\tiny$\pm$ 1.86} \\

    \noalign{\smallskip}
    \hline
    \noalign{\smallskip}
    
    \textbf{DCM} & \textbf{64}   
    & 13.75 {\tiny$\pm$ 2.13} 
    & 32.43 {\tiny$\pm$ 4.53} 
    & {35.14} {\tiny$\pm$ 2.56} 
    & 32.28 {\tiny$\pm$ 4.83} 
    & 57.92 {\tiny$\pm$ 4.59} 
    & {62.15} {\tiny$\pm$ 2.55} 
    & 57.15 {\tiny$\pm$ 2.20} 
    & 66.64 {\tiny$\pm$ 3.62} 
    & {69.80} {\tiny$\pm$ 3.38} \\
    \textbf{DCM} & \textbf{128}
    & 19.63 {\tiny$\pm$ 2.87} & {41.46} {\tiny$\pm$ 3.37} & 38.38 {\tiny$\pm$ 2.12} 
    & 39.11 {\tiny$\pm$ 3.06} & {64.38} {\tiny$\pm$ 1.89} & 63.96 {\tiny$\pm$ 1.44} 
    & 67.01 {\tiny$\pm$ 2.12} & 69.04 {\tiny$\pm$ 5.26} & {73.78} {\tiny$\pm$ 2.07} \\
    \textbf{DCM} & \textbf{256}   
    & 23.04 {\tiny$\pm$ 1.41} & 36.56 {\tiny$\pm$ 2.20} & {42.73} {\tiny$\pm$ 0.95} 
    & 49.00 {\tiny$\pm$ 4.13} & 62.80 {\tiny$\pm$ 3.14} & {68.25} {\tiny$\pm$ 1.39} 
    & 68.54 {\tiny$\pm$ 0.93} & {\underline{75.34}} {\tiny$\pm$ 1.30} & 73.72 {\tiny$\pm$ 2.35} \\
    \textbf{DCM} & \textbf{512}   
    & 29.13 {\tiny$\pm$ 2.90} & 38.04 {\tiny$\pm$ 2.73} & {43.93} {\tiny$\pm$ 1.65} 
    & 55.09 {\tiny$\pm$ 2.46} & 67.02 {\tiny$\pm$ 2.74} & {71.50} {\tiny$\pm$ 1.38} 
    & 71.76 {\tiny$\pm$ 2.48} & 71.01 {\tiny$\pm$ 2.95} & {74.85} {\tiny$\pm$ 0.28} \\
    \textbf{DCM} & \textbf{1024}   
    & 34.78 {\tiny$\pm$ 2.26} & 44.06 {\tiny$\pm$ 2.09} & {46.70} {\tiny$\pm$ 0.56} 
    & 62.53 {\tiny$\pm$ 2.23} & 71.73 {\tiny$\pm$ 1.69} & {73.17} {\tiny$\pm$ 0.67} 
    & 74.66 {\tiny$\pm$ 5.09} 
    & {\underline{78.06}} {\tiny$\pm$ 2.89} 
    & \underline{75.93} {\tiny$\pm$ 0.52} \\
    
    \textbf{DCM} & \textbf{All}
    & 45.01 {\tiny$\pm$ 1.67} & 47.22 {\tiny$\pm$ 1.55} & {49.24} {\tiny$\pm$ 0.22}	
    & 68.17 {\tiny$\pm$ 2.62} & 72.25 {\tiny$\pm$ 1.01} & {73.26} {\tiny$\pm$ 0.04}
    & \underline{78.27} {\tiny$\pm$ 0.32} & {\textbf{79.48}} {\tiny$\pm$ 3.48} & \underline{75.87} {\tiny$\pm$ 2.79} \\

    \noalign{\smallskip}
    \hline
    \noalign{\smallskip}

    \noalign{\smallskip}
    \hline
    \noalign{\smallskip}
    
    \textbf{Training} & \textbf{\# of} 
    & \multicolumn{3}{c|}{\textbf{M109-B}} 
    & \multicolumn{3}{c|}{\textbf{DCM-B}} 
    & \multicolumn{3}{c}{\textbf{C2k*}} \\
    \textbf{Datasets} & \textbf{images} 
    & \textbf{N} & \textbf{ST} & \textbf{SS}
    & \textbf{N} & \textbf{ST} & \textbf{SS}
    & \textbf{N} & \textbf{ST} & \textbf{SS} \\
    
    \noalign{\smallskip}
    \hline
    \noalign{\smallskip}

    \textbf{M109} & \textbf{64}   
    & 54.36 {\tiny$\pm$ 4.04} & 72.52 {\tiny$\pm$ 1.07} & 77.72 {\tiny$\pm$ 0.74}
    & 20.83 {\tiny$\pm$ 5.11} & 66.94 {\tiny$\pm$ 1.73} & 74.47 {\tiny$\pm$ 0.38} 
    & 26.41 {\tiny$\pm$ 6.28} & 55.98 {\tiny$\pm$ 3.38} & 62.57 {\tiny$\pm$ 1.61} \\
    
    \textbf{M109} & \textbf{128}
    & 64.18 {\tiny$\pm$ 2.18} & 75.58 {\tiny$\pm$ 1.25} & 79.27 {\tiny$\pm$ 0.74} 
    & 31.44 {\tiny$\pm$ 6.17} & 66.96 {\tiny$\pm$ 3.80} & 72.52 {\tiny$\pm$ 2.75} 
    & 36.91 {\tiny$\pm$ 4.41} & 58.29 {\tiny$\pm$ 1.69} & 62.59 {\tiny$\pm$ 1.91} \\
    
    \textbf{M109} & \textbf{256}   
    & 71.68 {\tiny$\pm$ 0.54} & 77.78 {\tiny$\pm$ 0.52} & \underline{80.79} {\tiny$\pm$ 0.61} 
    & 36.68 {\tiny$\pm$ 4.07} & 64.30 {\tiny$\pm$ 4.78} & 73.13 {\tiny$\pm$ 0.28} 
    & 44.22 {\tiny$\pm$ 2.35} & 59.66 {\tiny$\pm$ 1.71} & 63.08 {\tiny$\pm$ 1.10}\\
    
    \textbf{M109} & \textbf{512}   
    & 75.37 {\tiny$\pm$ 0.28} & \underline{81.63} {\tiny$\pm$ 1.25} & \underline{82.32} {\tiny$\pm$ 0.24} 
    & 47.61 {\tiny$\pm$ 2.15} & 69.46 {\tiny$\pm$ 3.27} & 74.87 {\tiny$\pm$ 1.20} 
    & 53.35 {\tiny$\pm$ 1.61} & 61.74 {\tiny$\pm$ 2.44} & 64.65 {\tiny$\pm$ 1.07} \\
    
    \textbf{M109} & \textbf{1024}   
    & \underline{79.67} {\tiny$\pm$ 1.48} & \underline{83.32} {\tiny$\pm$ 0.40} & \underline{83.51} {\tiny$\pm$ 0.35} 
    & 53.46 {\tiny$\pm$ 3.83} & 71.34 {\tiny$\pm$ 2.07} & \underline{76.89} {\tiny$\pm$ 0.68} 
    & 56.81 {\tiny$\pm$ 1.83} & 63.96 {\tiny$\pm$ 0.87} & 66.47 {\tiny$\pm$ 1.46} \\
    
    \textbf{M109} & \textbf{All}
    & \underline{85.78} {\tiny$\pm$ 0.22} & \underline{\textbf{87.15}} {\tiny$\pm$ 0.04} & \underline{87.06} {\tiny$\pm$ 0.10}	
    & 71.27 {\tiny$\pm$ 2.46} & 75.30 {\tiny$\pm$ 1.13} & \underline{78.06} {\tiny$\pm$ 1.12}
    & 66.39 {\tiny$\pm$ 1.21} & 69.24 {\tiny$\pm$ 0.93} & 70.49 {\tiny$\pm$ 0.68} \\

    \noalign{\smallskip}
    \hline
    \noalign{\smallskip}

    \textbf{DCM} & \textbf{64}   
    & 37.51 {\tiny$\pm$ 4.20} & 62.79 {\tiny$\pm$ 3.71} & 68.90 {\tiny$\pm$ 1.18} 
    & 46.55 {\tiny$\pm$ 1.78} & 71.97 {\tiny$\pm$ 4.72} & \underline{77.28} {\tiny$\pm$ 1.45} 
    & 31.60 {\tiny$\pm$ 2.95} & 58.19 {\tiny$\pm$ 3.87} & 64.40 {\tiny$\pm$ 1.03} \\
    
    \textbf{DCM} & \textbf{128}
    & 43.57 {\tiny$\pm$ 3.77} & 66.92 {\tiny$\pm$ 3.57} & 68.29 {\tiny$\pm$ 1.47} 
    & 54.80 {\tiny$\pm$ 1.04} & 76.71 {\tiny$\pm$ 1.94} & \underline{78.34} {\tiny$\pm$ 0.36} 
    & 36.67 {\tiny$\pm$ 1.93} & 60.99 {\tiny$\pm$ 4.36} & 65.13 {\tiny$\pm$ 0.73} \\
    
    \textbf{DCM} & \textbf{256}   
    & 53.20 {\tiny$\pm$ 1.41} & 66.17 {\tiny$\pm$ 2.76} & 71.40 {\tiny$\pm$ 1.45} 
    & 64.63 {\tiny$\pm$ 1.22} & \underline{79.27} {\tiny$\pm$ 1.03} & \underline{80.91} {\tiny$\pm$ 1.59} 
    & 43.77 {\tiny$\pm$ 2.64} & 60.05 {\tiny$\pm$ 1.96} & 66.10 {\tiny$\pm$ 0.49} \\
    
    \textbf{DCM} & \textbf{512}   
    & 58.32 {\tiny$\pm$ 4.07} & 70.29 {\tiny$\pm$ 2.42} & 73.60 {\tiny$\pm$ 0.47} 
    & 69.93 {\tiny$\pm$ 1.45} & \underline{80.84} {\tiny$\pm$ 1.44} & \underline{82.40} {\tiny$\pm$ 0.57} 
    & 49.96 {\tiny$\pm$ 2.33} & 62.78 {\tiny$\pm$ 2.10} & 67.38 {\tiny$\pm$ 0.67} \\
    
    \textbf{DCM} & \textbf{1024}   
    & 64.58 {\tiny$\pm$ 0.86} & 72.00 {\tiny$\pm$ 1.78} & 74.96 {\tiny$\pm$ 0.51} 
    & \underline{76.89} {\tiny$\pm$ 1.27} 
    & \underline{83.31} {\tiny$\pm$ 0.58} 
    & \underline{83.81} {\tiny$\pm$ 0.43} 
    & 56.75 {\tiny$\pm$ 1.51} & 64.57 {\tiny$\pm$ 1.72} & 68.00 {\tiny$\pm$ 0.69} \\

    \textbf{DCM} & \textbf{All}
    & 68.01 {\tiny$\pm$ 1.33} & {70.42} {\tiny$\pm$ 2.17} & 70.09 {\tiny$\pm$ 0.39}	
    & \underline{81.16} {\tiny$\pm$ 0.29} 
    & \underline{84.66} {\tiny$\pm$ 0.55} 
    & \underline{\textbf{84.89}} {\tiny$\pm$ 0.20}
    & 62.19 {\tiny$\pm$ 1.40} & 65.67 {\tiny$\pm$ 1.24} & 65.73 {\tiny$\pm$ 0.63} \\

    \noalign{\smallskip}
    \hline
    \noalign{\smallskip}
    
    \textbf{C2k*} & \textbf{64}   
    & 31.04 {\tiny$\pm$ 1.96} & 56.78 {\tiny$\pm$ 4.99} & 67.94 {\tiny$\pm$ 1.51} 
    & 24.20 {\tiny$\pm$ 3.34} & 58.17 {\tiny$\pm$ 3.11} & 72.52 {\tiny$\pm$ 0.84} 
    & 41.29 {\tiny$\pm$ 1.84} & 62.66 {\tiny$\pm$ 1.95} & 67.59 {\tiny$\pm$ 1.57} \\
    \textbf{C2k*} & \textbf{128}
    & 40.02 {\tiny$\pm$ 4.74} & 62.81 {\tiny$\pm$ 2.46} & 69.64 {\tiny$\pm$ 2.06} 
    & 32.06 {\tiny$\pm$ 5.57} & 65.39 {\tiny$\pm$ 3.61} & 71.86 {\tiny$\pm$ 2.24} 
    & 49.83 {\tiny$\pm$ 3.15} & 66.69 {\tiny$\pm$ 1.38} & 69.45 {\tiny$\pm$ 1.27} \\
    \textbf{C2k*} & \textbf{256}   
    & 50.43 {\tiny$\pm$ 3.75} & 67.76 {\tiny$\pm$ 2.48} & 72.13 {\tiny$\pm$ 0.74} 
    & 45.63 {\tiny$\pm$ 1.50} & 67.12 {\tiny$\pm$ 4.00} & 73.43 {\tiny$\pm$ 1.48} 
    & 57.59 {\tiny$\pm$ 2.02} & 69.21 {\tiny$\pm$ 0.87} & 70.55 {\tiny$\pm$ 1.32} \\
    \textbf{C2k*} & \textbf{512}   
    & 58.60 {\tiny$\pm$ 0.52} & 67.87 {\tiny$\pm$ 1.52} & 72.53 {\tiny$\pm$ 1.71} 
    & 54.99 {\tiny$\pm$ 2.37} & 69.26 {\tiny$\pm$ 4.94} & 72.92 {\tiny$\pm$ 1.24} 
    & 64.98 {\tiny$\pm$ 1.77} & 70.90 {\tiny$\pm$ 0.93} & 73.82 {\tiny$\pm$ 1.12} \\
    \textbf{C2k*} & \textbf{1024}   
    & 63.20 {\tiny$\pm$ 1.60} & 70.78 {\tiny$\pm$ 0.98} & 75.87 {\tiny$\pm$ 0.41} 
    & 59.03 {\tiny$\pm$ 0.92} & 71.03 {\tiny$\pm$ 1.44} & \underline{76.84} {\tiny$\pm$ 0.65} 
    & 73.35 {\tiny$\pm$ 1.00} & \underline{76.19} {\tiny$\pm$ 1.41} & \underline{76.34} {\tiny$\pm$ 0.55} \\
    
    \textbf{C2k*} & \textbf{All}
    & 68.13 {\tiny$\pm$ 1.84} & 71.81 {\tiny$\pm$ 0.35} & 71.57 {\tiny$\pm$ 0.19}	
    & 67.78 {\tiny$\pm$ 1.99} & 70.29 {\tiny$\pm$ 0.67} & 71.10 {\tiny$\pm$ 0.52}
    & \underline{76.19} {\tiny$\pm$ 1.32} & \underline{79.90} {\tiny$\pm$ 0.61} & \underline{\textbf{79.93}} {\tiny$\pm$ 0.28} \\

    \noalign{\smallskip}
    \hline
\end{tabular}
\caption{AP performances of our model after stage 3 fine-tuning when trained with a subset of individual datasets having annotations of a limited number of random images. c2k* indicates that all Comic2k, Watercolor2k, and Clipart1k datasets are combined for training the model. N: no pre-training, ST: pre-trained with style transferred images, SS: additional teacher-student pre-training. \underline{Underlined} if the score of the evaluated dataset is higher than the previous supervised SOTA detector, and \textbf{bold} if the score is the best for the particular dataset in this table.}
\label{table:sup_stage_3_results}
\end{table*}

%% file: ijcai23.bbl
\begin{thebibliography}{}

\bibitem[\protect\citeauthoryear{Bochkovskiy \bgroup \em et al.\egroup
  }{2020}]{bochkovskiy2020yolov4}
Alexey Bochkovskiy, Chien-Yao Wang, and Hong-Yuan~Mark Liao.
\newblock Yolov4: Optimal speed and accuracy of object detection, 2020.

\bibitem[\protect\citeauthoryear{Brumm \bgroup \em et al.\egroup
  }{2021}]{brumm2021oldest}
Adam Brumm, Adhi~Agus Oktaviana, Basran Burhan, Budianto Hakim, Rustan Lebe,
  Jian-xin Zhao, Priyatno~Hadi Sulistyarto, Marlon Ririmasse, Shinatria
  Adhityatama, Iwan Sumantri, et~al.
\newblock Oldest cave art found in sulawesi.
\newblock {\em Science Advances}, 7(3):eabd4648, 2021.

\bibitem[\protect\citeauthoryear{Cai \bgroup \em et al.\egroup
  }{2019}]{mtor_model}
Qi~Cai, Yingwei Pan, Chong-Wah Ngo, Xinmei Tian, Lingyu Duan, and Ting Yao.
\newblock Exploring object relation in mean teacher for cross-domain detection,
  2019.

\bibitem[\protect\citeauthoryear{Chen \bgroup \em et al.\egroup
  }{2018}]{chen2018cartoongan}
Yang Chen, Yu-Kun Lai, and Yong-Jin Liu.
\newblock Cartoongan: Generative adversarial networks for photo cartoonization.
\newblock In {\em Proceedings of the IEEE conference on computer vision and
  pattern recognition}, pages 9465--9474, 2018.

\bibitem[\protect\citeauthoryear{Deng \bgroup \em et al.\egroup
  }{2020}]{umt_model}
Jinhong Deng, Wen Li, Yuhua Chen, and Lixin Duan.
\newblock Unbiased mean teacher for cross-domain object detection, 2020.

\bibitem[\protect\citeauthoryear{Ge \bgroup \em et al.\egroup
  }{2021}]{yolox2021}
Zheng Ge, Songtao Liu, Feng Wang, Zeming Li, and Jian Sun.
\newblock Yolox: Exceeding yolo series in 2021.
\newblock {\em arXiv preprint arXiv:2107.08430}, 2021.

\bibitem[\protect\citeauthoryear{Hicsonmez \bgroup \em et al.\egroup
  }{2020}]{DBLP:journals/corr/abs-2002-05638}
Samet Hicsonmez, Nermin Samet, Emre Akbas, and Pinar Duygulu.
\newblock {GANILLA:} generative adversarial networks for image to illustration
  translation.
\newblock {\em CoRR}, abs/2002.05638, 2020.

\bibitem[\protect\citeauthoryear{Inoue \bgroup \em et al.\egroup
  }{2018}]{inoue2018crossdomain}
Naoto Inoue, Ryosuke Furuta, Toshihiko Yamasaki, and Kiyoharu Aizawa.
\newblock Cross-domain weakly-supervised object detection through progressive
  domain adaptation, 2018.

\bibitem[\protect\citeauthoryear{Iyyer \bgroup \em et al.\egroup
  }{2017}]{iyyer2017amazing}
Mohit Iyyer, Varun Manjunatha, Anupam Guha, Yogarshi Vyas, Jordan Boyd-Graber,
  Hal Daumé~III au2, and Larry Davis.
\newblock The amazing mysteries of the gutter: Drawing inferences between
  panels in comic book narratives, 2017.

\bibitem[\protect\citeauthoryear{Jiang \bgroup \em et al.\egroup
  }{2022}]{dadapt}
Junguang Jiang, Baixu Chen, Jianmin Wang, and Mingsheng Long.
\newblock Decoupled adaptation for cross-domain object detection.
\newblock In {\em International Conference on Learning Representations}, 2022.

\bibitem[\protect\citeauthoryear{Lin \bgroup \em et al.\egroup
  }{2014}]{lin2014microsoft}
Tsung-Yi Lin, Michael Maire, Serge Belongie, James Hays, Pietro Perona, Deva
  Ramanan, Piotr Doll{\'a}r, and C~Lawrence Zitnick.
\newblock Microsoft coco: Common objects in context.
\newblock In {\em European conference on computer vision}, pages 740--755.
  Springer, 2014.

\bibitem[\protect\citeauthoryear{Liu \bgroup \em et al.\egroup
  }{2021}]{liu2021unbiased}
Yen-Cheng Liu, Chih-Yao Ma, Zijian He, Chia-Wen Kuo, Kan Chen, Peizhao Zhang,
  Bichen Wu, Zsolt Kira, and Peter Vajda.
\newblock Unbiased teacher for semi-supervised object detection, 2021.

\bibitem[\protect\citeauthoryear{Matsui \bgroup \em et al.\egroup
  }{2017}]{mtap_matsui_2017}
Yusuke Matsui, Kota Ito, Yuji Aramaki, Azuma Fujimoto, Toru Ogawa, Toshihiko
  Yamasaki, and Kiyoharu Aizawa.
\newblock Sketch-based manga retrieval using manga109 dataset.
\newblock {\em Multimedia Tools and Applications}, 76(20):21811--21838, 2017.

\bibitem[\protect\citeauthoryear{Nguyen \bgroup \em et al.\egroup
  }{2018}]{jimaging4070089}
Nhu-Van Nguyen, Christophe Rigaud, and Jean-Christophe Burie.
\newblock Digital comics image indexing based on deep learning.
\newblock {\em Journal of Imaging}, 4(7), 2018.

\bibitem[\protect\citeauthoryear{Ogawa \bgroup \em et al.\egroup
  }{2018}]{ogawa2018object}
Toru Ogawa, Atsushi Otsubo, Rei Narita, Yusuke Matsui, Toshihiko Yamasaki, and
  Kiyoharu Aizawa.
\newblock Object detection for comics using manga109 annotations, 2018.

\bibitem[\protect\citeauthoryear{Shrivastava \bgroup \em et al.\egroup
  }{2016}]{shrivastava2016training}
Abhinav Shrivastava, Abhinav Gupta, and Ross Girshick.
\newblock Training region-based object detectors with online hard example
  mining, 2016.

\bibitem[\protect\citeauthoryear{Wang and Yu}{2020}]{Wang_2020_CVPR}
Xinrui Wang and Jinze Yu.
\newblock Learning to cartoonize using white-box cartoon representations.
\newblock In {\em IEEE/CVF Conference on Computer Vision and Pattern
  Recognition (CVPR)}, June 2020.

\bibitem[\protect\citeauthoryear{Xu \bgroup \em et al.\egroup
  }{2021}]{xu2021end}
Mengde Xu, Zheng Zhang, Han Hu, Jianfeng Wang, Lijuan Wang, Fangyun Wei, Xiang
  Bai, and Zicheng Liu.
\newblock End-to-end semi-supervised object detection with soft teacher.
\newblock {\em Proceedings of the IEEE/CVF International Conference on Computer
  Vision (ICCV)}, 2021.

\bibitem[\protect\citeauthoryear{Xu \bgroup \em et al.\egroup
  }{2022}]{h2farcnn}
Yunqiu Xu, Yifan Sun, Zongxin Yang, Jiaxu Miao, and Yi~Yang.
\newblock {H$^2$FA R-CNN}: Holistic and hierarchical feature alignment for
  cross-domain weakly supervised object detection.
\newblock In {\em Proceedings of IEEE/CVF Conference on Computer Vision and
  Pattern Recognition (CVPR)}, pages 14329--14339, 2022.

\bibitem[\protect\citeauthoryear{Yang \bgroup \em et al.\egroup
  }{2016}]{yang2016wider}
Shuo Yang, Ping Luo, Chen~Change Loy, and Xiaoou Tang.
\newblock Wider face: A face detection benchmark.
\newblock In {\em IEEE Conference on Computer Vision and Pattern Recognition
  (CVPR)}, 2016.

\bibitem[\protect\citeauthoryear{Zhang \bgroup \em et al.\egroup
  }{2020a}]{zhang2020acfd}
Bin Zhang, Jian Li, Yabiao Wang, Zhipeng Cui, Yili Xia, Chengjie Wang, Jilin
  Li, and Feiyue Huang.
\newblock Acfd: Asymmetric cartoon face detector, 2020.

\bibitem[\protect\citeauthoryear{Zhang \bgroup \em et al.\egroup
  }{2020b}]{DynamicRCNN}
Hongkai Zhang, Hong Chang, Bingpeng Ma, Naiyan Wang, and Xilin Chen.
\newblock Dynamic {R-CNN}: Towards high quality object detection via dynamic
  training.
\newblock In {\em ECCV}, 2020.

\bibitem[\protect\citeauthoryear{Zheng \bgroup \em et al.\egroup
  }{2020}]{zheng2020cartoon}
Yi~Zheng, Yifan Zhao, Mengyuan Ren, He~Yan, Xiangju Lu, Junhui Liu, and Jia Li.
\newblock Cartoon face recognition: A benchmark dataset.
\newblock In {\em Proceedings of the 28th ACM International Conference on
  Multimedia}, pages 2264--2272, 2020.

\bibitem[\protect\citeauthoryear{Zhu \bgroup \em et al.\egroup
  }{2017}]{CycleGAN2017}
Jun-Yan Zhu, Taesung Park, Phillip Isola, and Alexei~A Efros.
\newblock Unpaired image-to-image translation using cycle-consistent
  adversarial networks.
\newblock In {\em Computer Vision (ICCV), 2017 IEEE International Conference
  on}, 2017.

\end{thebibliography}
